# Performance evaluation of predictive AI models to support medical decisions: Overview and guidance

Ben VAN CALSTER[1,2,3], Gary S COLLINS[4], Andrew J VICKERS[5], Laure WYNANTS[1,2,6], Kathleen F KERR[7], Lasai BARREÑADA[1,3], Gael VAROQUAUX[8], Karandeep SINGH[9], Karel GM MOONS[10], Tina HERNANDEZ-BOUSSARD[11,12], Dirk TIMMERMAN[1,13], David J McLERNON[14], Maarten VAN SMEDEN[10], Ewout W STEYERBERG[3,10], for topic group 6 of the STRATOS initiative

1 Dept Development and Regeneration, KU Leuven, Leuven, Belgium

2 Leuven Unit for Health Technology Assessment Research (LUHTAR), KU Leuven, Leuven, Belgium

3 Dept Biomedical Data Sciences, Leiden University Medical Center, Leiden, the Netherlands

4 Centre for Statistics in Medicine, Nuffield Department of Orthopaedics, Rheumatology, and Musculoskeletal Sciences, University of Oxford, UK

5 Department of Epidemiology and Biostatistics, Memorial Sloan Kettering Cancer Center, New York, NY, USA

6 Department of Epidemiology, CAPHRI Care and Public Health Research Institute, Maastricht University, Maastricht, Netherlands

7 Department of Biostatistics, University of Washington School of Public Health, Seattle, WA, USA

8 Parietal project team, INRIA Saclay-Île de France, Palaiseau, France

9 Division of Biomedical Informatics, Department of Medicine, University of California, San Diego, CA, USA

10 Julius Centre for Health Sciences and Primary Care, University Medical Centre Utrecht, Utrecht University, Utrecht, Netherlands

11 Department of Medicine (Biomedical Informatics), Stanford University, Stanford, CA, USA

12 Department of Biomedical Data Science, Stanford University, Stanford, CA, USA

13 Department of Obstetrics and Gynecology, University Hospitals Leuven, Leuven, Belgium

14 Institute of Applied Health Sciences, University of Aberdeen, Aberdeen, UK

**Word count main text: 6927**

**Abstract**

A myriad of measures to illustrate performance of predictive artificial intelligence (AI) models have been proposed in the literature. Selecting appropriate performance measures is essential for predictive AI models that are developed to be used in medical practice, because poorly performing models may harm patients and lead to increased costs. We aim to assess the merits of classic and contemporary performance measures when validating predictive AI models for use in medical practice. We focus on models with a binary outcome.

We discuss 32 performance measures covering five performance domains (discrimination, calibration, overall, classification, and clinical utility) along with accompanying graphical assessments. The first four domains cover statistical performance, the fifth domain covers decision-analytic performance. We explain why two key characteristics are important when selecting which performance measures to assess: (1) whether the measure's expected value is optimized when it is calculated using the correct probabilities (i.e., a "proper" measure), and (2) whether they reflect either purely statistical performance or decision-analytic performance by properly considering misclassification costs. Seventeen measures exhibit both characteristics, fourteen measures exhibited one characteristic, and one measure possessed neither characteristic (the F1 measure). All classification measures (such as classification accuracy and F1) are improper for clinically relevant decision thresholds other than 0.5 or the prevalence.

We recommend the following measures and plots as essential to report: AUROC, calibration plot, a clinical utility measure such as net benefit with decision curve analysis, and a plot with probability distributions per outcome category.

## 1. Introduction

The medical literature abounds with predictive artificial intelligence (AI) models that estimate the probability of individuals having (diagnostic) or developing (prognostic) an disease or health state of interest (the 'event'), also known as clinical prediction models.[1] Traditionally, these models were developed using statistical methods such as regression analysis, but more flexible machine learning algorithms are being increasingly used. For instance, a traditional logistic regression model aims to predict the risk of a permanent stoma in patients undergoing a resection of left-sided obstructive colon cancer using demographic, clinical, and laboratory measurements.[2] A more contemporary model built using deep learning aims to predict the presence of atrial fibrillation based on sinus rhythm electrocardiograms.[3]

Regardless of the modeling approach, it is essential to properly evaluate the performance of predictive AI models that were developed with the intention of being used in medical practice. Numerous measures have been suggested, but there is lack of clarity and occasional conflict regarding which measures are recommended in the medical, statistical, and machine learning literature.[4-11] Selecting appropriate performance measures for predictive AI in healthcare is essential, in particular when the intention is to deploy models in patient care settings. If models do not perform well, they may harm patients and lead to increased costs.[12] Performance assessment is primarily important for external validation studies, but also for internal validation studies. An external validation study evaluates the performance of a model in a dataset that includes individual participant data from a target population where the model might be employed in medical practice.[13-16] Compared to the development (or training) dataset, the validation dataset includes individuals from a different location, time period, or setting. In contrast, internal validation evaluates model performance in data from the same population as the development dataset using methods such as cross-validation, bootstrapping, or (repeated) train-test splitting.[13,14] Internal validation does not refer to model selection, but to an independent evaluation of the selected model.

The aims of this paper are to assess classic and contemporary performance measures from the statistical and machine learning literature, and provide recommendations for researchers, end-users and stakeholders. We present a taxonomy of performance domains, describe key characteristics for performance measures, discuss performance measures in combination with an illustrative case study, and provide guidance. We target models that predict a binary outcome by estimating the probability of the event, although the general principles generalize to models for predicting nominal, ordinal, continuous, or time-to-event outcomes that include censored observations.

Ultimately, we recommend the following measures and plots to be used in most circumstances: AUROC, calibration plot, a clinical utility measure such as net benefit with decision curve analysis, and a plot with probability distributions per outcome category.

## 2. Case study: external validation of diagnostic model for ovarian cancer

We consider prediction of malignancy in women with an ovarian tumor. The ADNEX model preoperatively estimates the probability of malignancy in women with an ovarian tumor that are scheduled for surgery.[17] The model can be used to make decisions regarding the type of surgery (advanced versus conservative) for patients examined at an oncology center, or regarding referral to an oncology center for patients examined elsewhere.[18] ADNEX was developed on data from 5909 individuals recruited between 1999 and 2012. The transIOTA study externally validated ADNEX to distinguish benign from malignant tumors, using data from 894 women recruited between 2015 and 2019, of whom 434 had a malignant tumor (prevalence 49%).[18] For didactic purposes, we use this dataset to calculate all discussed performance measures with 95% confidence intervals and demonstrate all discussed visualizations. Confidence intervals were obtained using the percentile bootstrap method on 1000 bootstrap samples. All R and Python scripts, as well as the estimated risk of malignancy and outcome of the 894 participants are provided on https://github.com/benvancalster/PerfMeasuresOverview.

## 3. A taxonomy of five performance domains

Three domains evaluate performance based on the probability estimates (Table 1). *Discrimination* performance focuses on the extent to which the model gives higher probability estimates for individuals with the event compared to those without the event. *Calibration* performance focuses on the extent to which the probability estimates correspond to observed event proportions. Models can have good discrimination but poor calibration, and vice versa. The *overall* predictive performance of a model captures elements of both discrimination and calibration by quantifying how closely the probability estimates approach the actual outcomes of 0 (no event) or 1 (event).[8,19,20]

When a single probability threshold is defined, individuals can be classified into two mutually exclusive groups which should be linked to a decision about an intervention (e.g., refer for a specific treatment, request additional tests). The general idea is that individuals with an estimated probability equal to or higher than the threshold are eligible for a specific intervention (they are at "high risk" for the event). The intervention would not be suggested for individuals with an estimated probability below the threshold ("low risk"). This is referred to as the 'decision threshold'. The decision threshold should be clinically relevant, in the sense that it should be linked with misclassification costs (Box 1). It is also possible to use multiple decision thresholds to separate individuals into three or more groups, however we will focus on the case where a single threshold is used. The fourth performance domain, *Classification*, focuses on the extent to which individuals are correctly classified as high or low risk. Classification performance is affected by the discrimination and calibration performance of the model. In biomedical applications, the consequences of a false negative (for instance, not referring a woman with ovarian malignancy for advanced surgery) are almost always different than those of a false positive (referring a woman without ovarian malignancy for advanced surgery). Typically, the former error is more severe than the latter. The fifth performance domain, *clinical utility*, incorporates such misclassification costs to determine whether use of the model leads to better decisions on average. While 'misclassification costs' is an established term, it refers broadly to any harms of misclassification (false positives and false negatives).[21,22]

We discuss 32 performance measures (3 discrimination, 6 calibration, 9 overall, 11 classification, and 3 clinical utility) (Table 2), along with accompanying visual assessments.

**Box 1. Defining a decision threshold**

The primary aim of most predictive AI models in medical practice is to support subsequent decision-making. Probability estimates may guide health professionals and patients to improve health outcomes by avoiding a burdensome intervention with limited expected benefit for those at low risk, and choosing the intervention for those at high risk. It follows naturally that the decision threshold should be defined based on medical rather than statistical arguments[23]. Searching for the threshold that optimizes a statistical measure such as the Youden index (sensitivity + specificity – 1) is inconsistent with decision theory and detached from practical usage by clinicians. Maximizing the Youden index assumes that sensitivity and specificity are equally important, which is almost never the case in medicine, and implicitly, that the optimal threshold is the prevalence. Instead, once the decision that the model intends to support is clearly defined, the four possible consequences of using the model to support that decision should be considered: true positives (individuals with the event that are classified as at high risk), true negatives (individuals without the event that are classified as at low risk), false negatives (individuals with the event that are classified as at low risk), and false positives (individuals without the event that are classified as at high risk).[21,24,25] The weight of these consequences may vary by patient characteristics, patient preference, healthcare system and even by clinician.

In the case study, when patients are scheduled for surgery, the ADNEX model can be used to decide whether advanced versus conservative surgery is needed. A decision threshold of 10% for the probability of malignancy is often recommended.[26] This threshold corresponds to performing advanced surgery in up to 10 individuals per true positive (i.e., performing advanced surgery in a patient with a malignant tumor). In other words, up to 9 false positives (i.e., performing advanced surgery in a patient with a benign tumor) are accepted per true positive (advanced surgery in a patient with malignancy). The medical benefit of operating on a malignant tumor (compared to not operating on a malignant tumor) is 9 times greater than the harm of unnecessary surgery in individuals with a benign tumor. A range of possible decision thresholds can be defined in consultation with medical experts.[27,28] For ADNEX, discussions with clinicians from different countries resulted in a range of possible decision thresholds that vary from 5% to 40%. Hence, we assume that probabilities below 5% would always be considered too low to recommend advanced surgery, while probabilities over 40% would always be too high to recommend a conservative approach. A decision threshold of 5% corresponds to accepting up to 19 false positives per true positive (odds 1:19), a decision threshold of 40% corresponds to accepting up to 3 false positives per 2 true positives (odds 2:3). Despite the variation in decision thresholds, the range is often below the prevalence (i.e., proportion of malignancy in the case study, 49%) if the default decision is to intervene (i.e., advanced surgery) and above the prevalence if the default decision is not to intervene (i.e., conservative surgery).[29] The default decision is the decision one would take in the absence of any predictive information. We write that the range is 'often' below the prevalence if the default decision is to intervene, because clinical reality is complicated. For example, the prevalence varies between settings such as hospitals, and resource constraints may play a role as well.[30]

## 4. Key characteristics of an informative performance measure

We define two key characteristics that a performance measure should meet: (1) measures should be proper, and (2) they should have a clear focus in reflecting either purely statistical value or decision-analytic value by properly considering differential misclassification costs as discussed in Box 1. A third desirable characteristic is an intuitive interpretation.[31,32] We do not discuss this further, as interpretability is subjective, influenced by background knowledge and familiarity.

*Properness*. A performance measure is called proper if its expected value is optimized when using the correct probabilities.[33-37] Other expressions for 'using the correct probabilities' are 'using the correct model', 'using the true probabilities', or 'using a strongly calibrated model'. A proper measure cannot be fooled: in expectation, the correct model cannot be beaten by an incorrect one. A measure is strictly proper when its expected value is only maximized when the correct probabilities are used. When the expected value is not uniquely maximized when using the correct probabilities, a measure is called semi-proper. When the expected value is not maximized when using the correct probabilities, the measure is termed improper. Table 2, lists the properness status for the 32 measures, Supplementary Appendix 1 provides an illustration.

*Clear focus on statistical or decision-analytic evaluation*. There is a clear distinction between statistical and decision-analytic performance evaluation of predictive AI for medical practice. Statistical performance measures have an essential role in evaluating models but they cannot be used to determine whether a model should be used in practice: it is not appropriate to cite, say, good discrimination and calibration and conclude "our model can be used to aid decisions about ovarian surgery". If a performance measure aims to go beyond measuring statistical value, then it needs to incorporate misclassification costs in accordance with decision-analytic principles. Calibration, discrimination, overall, and classification performance domains are limited to aspects of statistical performance. The clinical utility domain explicitly assesses value of the model to support decision-making.

## 5. Probability-based evaluation

### 5.1. Discrimination

The definition of discrimination implies that discrimination measures should only rely on the ranks of the estimated probabilities in the dataset.[38] The concordance probability (c statistic) is the proportion of pairs of individuals, consisting of one with the event and one without, for which the estimated probabilities are concordant with the outcome (Table S1-S2 for details and formulas). Concordance means that the estimated probability is higher for the individual with event than for the individual without. The c statistic is interpreted as the probability that the model can correctly discriminate a random individual with an event from a random individual without. For binary outcomes, the c statistic is equal to the area under the receiver operating characteristic curve (often referred to as AUC or AUROC) and to the Mann–Whitney U test statistic divided by the number of discordant pairs (i.e., the number of individuals with an event multiplied by the number of individuals without an event).[39,40] The ROC curve plots sensitivity against 1-specificity for all possible probability thresholds (Figure 1A for the case study).[41,42] The AUROC is 0.5 for a model with no discriminatory ability, and 1 for a model with perfect discrimination.

Several researchers have advised against the use of AUROC when prevalence is far from 0.5 (‘class imbalance’). The argument is that AUROC is misleading when the event is rare because (1) it ignores the difficulty of obtaining both acceptable positive predictive value (PPV or ‘precision’) and sensitivity (or ‘recall’), or (2) it does not consider misclassification costs.[43-48] PPV refers to the proportion of individuals with the event among those at high risk, sensitivity to the proportion of individuals at high risk among those with the event. There are no grounds to call AUROC misleading.[49] Discrimination measures are not meant to reflect differential misclassification costs, and class imbalance should not be conflated with misclassification costs or medical relevance. AUROC has a clear interpretation, is semi-proper, and assesses discrimination by quantifying to what extent the estimated probabilities separate individuals with and without an event.[38] While discriminatory ability is essential for predictive AI, we agree that the AUROC does not tell the full story. For example, as discrimination does not focus on misclassification costs, there are no AUROC values above which a model can be labeled as ‘useful’.[50] This holds both for balanced and imbalanced outcomes.

Nevertheless, the precision-recall curve and the area underneath (AUPRC) are often recommended as alternatives to the ROC curve and AUROC under class imbalance, claiming that AUROC overestimates performance.[45,51] The precision-recall curve plots precision by recall for all possible probability thresholds (Figure 1B for the case study). Better models have a curve closer to the upper right corner. Importantly, focusing on precision and recall implies that true negatives are not directly considered. Whereas true negatives may be largely irrelevant for specific non-medical applications, this is generally not the case for medical applications and is not in line with decision-analytic principles. The AUPRC has no clear interpretation and depends on the prevalence, which goes beyond assessing discrimination. AUPRC tends to give lower values than AUROC under class imbalance, not because AUROC overestimates performance but because AUPRC and AUROC look at different things.  A final technical comment is that the calculation of AUPRC is not straightforward.[48,52] The average precision (AP) is used as a simple measure to approximate the AUPRC.[52]

Another measured suggested as an alternative for AUROC is the partial AUROC (pAUROC).[46,53] pAUROC focuses on the part of ROC curve where the specificity or sensitivity reach a certain minimum tolerable level when using the model for decision-making.  However, a minimum tolerable specificity (c.q. sensitivity) depends on prevalence and sensitivity (c.q. specificity).[54] A statement such as “we need at least 90% sensitivity because we want to find at least 90% of the cancer cases" seems reasonable at face value. Depending on specificity and prevalence this could require very different decision thresholds to classify individuals at high risk. Consequently, there is no decision-analytic basis for this approach. The intuitive interpretation of the AUROC is lost by considering only a portion of the curve. The interpretation is the average sensitivity (c.q. specificity) over the range of tolerable specificity (c.q. sensitivity) levels. Figure 1C visualizes pAUROC for the case study, if we make the (insupportable) argument that a sensitivity below 0.8 is unacceptably low.

AUROC, AUPRC, and pAUROC are not strictly proper, because these rank-based measures are invariant to monotonic transformations of probability estimates. For instance, if we divided the probabilities of the ADNEX model by 100, so that even a woman at extremely high risk of cancer would be advised against advanced surgery, AUROC, AUPRC and pAUROC would not change. Critically, however, AUPRC and pAUROC conflate discrimination and clinical utility: ignoring true negative (AUPRC) or limiting the acceptable values for specificity or sensitivity (pAUROC) indirectly refers to misclassification costs. While showing the ROC or precision-recall curve is not wrong, in our experience these plots do not provide useful additional information beyond that provided by the summary measures (e.g. AUROC) and appropriate clinical utility measures.[41,49]

The ADNEX model had an AUROC of 0.91 (95% confidence interval 0.89-0.93), and an AUPRC of 0.89 (0.86-0.91) (Table 3). Ignoring sensitivity values below 0.8, pAUROC was 0.14 (0.13-0.15).

### 5.2. Calibration

Several tools to address calibration have been suggested in the statistical and machine learning literature. These tools can be classified into four increasingly stringent levels that have been labeled as mean, weak, moderate, and strong calibration.[55] The first two levels are mostly known from the statistical literature. We discuss performance measures for each level (Table S1-S2).

#### 5.2.1. Mean calibration (“calibration-in-the-large”)

At the simplest level, one can evaluate whether the model’s average estimated probability equals the observed prevalence. Two measures for calibration-in-the-large are the observed over expected ratio (O:E ratio) and the calibration intercept. The O:E ratio is the ratio of the total observed and total expected number of individuals with an event, where the model's expected number equals the sum of the estimated probabilities in the dataset. An O:E ratio > 1 indicates underestimation, whilst an O:E ratio < 1 indicates overestimation of the event probability. ADNEX had an O:E ratio of 1.23 (95% confidence interval 1.17-1.29), meaning that there were 23% more events observed than expected based on the model (Table 3).

The calibration Intercept refers to the intercept of a recalibration model. The recalibration model is a logistic model that has the logit of the estimated probability (the 'linear predictor') as the only covariate for which its coefficient is fixed to 1 by using the linear predictor as an offset term. A value > 0 indicates underestimation, whilst a value < 0 indicates overestimation. ADNEX had a calibration intercept of 0.81 (95% confidence interval 0.62-1.01), suggesting underestimation of probabilities on average. The O:E ratio has a more intuitive interpretation than the calibration intercept.

#### 5.2.2. Calibration in the weak sense

A model is calibrated in the weak sense if calibration-in-the-large is satisfied and the estimated probabilities have on average neither too much nor too little spread. Estimated probabilities with too much spread are on average too close to 0 and 1 (i.e., too confident), whilst estimated probabilities with too little spread are on average too close to the prevalence (i.e., too modest).[55,56] This spread can be quantified by the calibration slope, which is the coefficient of the linear predictor in a calibration model with the linear predictor as the only covariate (i.e., not constrained to 1 as done in the calibration model for the calibration-in-the-large).[57] A calibration slope < 1 indicates that probabilities are overestimated for those at highest risk and underestimated for those at lowest risk (too much spread), a calibration slope > 1 indicates the opposite (too little spread). To evaluate weak calibration, both the calibration intercept (or O:E ratio) and calibration slope need to be assessed. These concepts are orthogonal dimensions, looking at 'height' (probabilities too high/low, or 'calibration-in-the-large') versus 'width' (probabilities too wide/narrow, or 'calibration-in-the-small'). ADNEX had a calibration slope of 0.93 (95% confidence interval 0.83-1.05), suggesting that the spread of the probabilities was about right.

#### 5.2.3. Calibration in the moderate sense

Moderate calibration means that, among people with an estimated probability of x, the observed proportion of the event also equals x. The most common way of assessing calibration in the moderate sense is through a calibration plot, also referred to as a reliability diagram.[58-60] Calibration plots can be based on grouping individuals or smoothing. Using grouping, individuals are grouped based on their estimated probabilities, and the plot presents the proportion of events by the average estimated probability for each group. Typically, 5 or 10 groups are created to be of equal size defined by percentiles of the estimated probabilities. The visual impression in the plot depends on how groups are defined. Also, individuals with very different estimated probabilities may still end up in the same group, such that grouping cannot comprehensively assess calibration in the moderate sense. Alternatively, the relation between the estimated probability and the proportion of events can be visualized by smoothing the regression of the outcome on the (logit of the) individual probabilities using loess, splines or polynomial regression.[59,61] Alternative smoothing methods exist, for example based on kernel density estimation.[62] Figure 2 presents a calibration plot using both grouping (10 groups of equal size based on deciles of estimated probability) and smoothing (loess) for the case study. The curve is almost entirely above the diagonal, suggesting that estimated probabilities across the whole range were underestimated in this population.

Several summary measures for the calibration plot have been suggested. The Expected Calibration Error (ECE) is a measure based on grouping.[63] ECE is the sum of the absolute differences between the average estimated probability and the observed proportion, weighted by the sample size of each group. For smoothed plots, the Estimated Calibration Index (ECI) and Integrated Calibration Index (ICI) have been proposed.[64-66] ICI is a smoothed version of ECE: it is the average of the absolute differences between estimated probability and observed proportion over individuals in the dataset. ECI averages squared differences instead, and it has been suggested to divide the value by the ECI of a null model that estimates the prevalence for all individuals.[64] Whereas ECI and ICI depend on the smoother, ECE depends on the grouping approach.[67] Statistical tests have also been proposed.[61,68,69] An outdated approach is to use the Hosmer-Lemeshow test for calibration assessment. The test focuses on the null hypothesis that the grouped data lie on the diagonal line. Statistical tests of calibration will often mislead because a non-significant result is sometimes used unjustly to claim good calibration (e.g., in small samples) and a statistically significant result might represent trivial miscalibration in a very large study.[69,70] Moreover, both the test statistic and the p-value do not provide insight into the direction and magnitude of miscalibration.[60,71] Furthermore, a single model can both underestimate and overestimate risks across different ranges of estimated probabilities, which is a phenomenon easily observed on plots but not in summary measures of calibration. Therefore, calibration plots including confidence intervals are the key tool for assessing calibration.

#### 5.2.4. Calibration in the strong sense

When a model is calibrated in the strong sense, estimated probabilities correspond to observed proportions for every combination of predictor values.[55,72] This essentially implies that the model is fully correct conditional on the included predictors. This Achieving calibration in the strong sense is practically unachievable.[55] Intermediate approaches can be used, that aim to evaluate calibration beyond the moderate level. One example is to assess calibration plots for various subgroups based on variables that may or may not be used as predictors in the model, among others to assess fairness of the model across prespecified minority or sensitive groups.[73] Figure S1 presents calibration plots by menopausal status for the case study. Overall measures based on loglikelihood or Brier score are sensitive to deviations from strong calibration, as are clinical utility measures. The difference between moderate and strong calibration has been referred to as 'grouping loss' in the machine learning literature, and current research focuses on its quantification.[74] More generally, what we call strong calibration is related to the notion of epistemic uncertainty.[75]

### 5.3. Overall performance

The (negative) loglikelihood is a central quantity that is often used to optimize the fit of models such as logistic regression or deep neural networks. The negative loglikelihood is sometimes used to assess overall performance, and in the machine learning literature, it is commonly referred to as logloss or cross-entropy (Table S1-S2).[76,77] The Brier score is calculated as the average squared difference between estimated probability of the event and the observed outcome.[78] The lower the Brier score, the closer the estimated probabilities approach the observed outcomes, with a value of 0 corresponding to a (utopic) perfect model that gives probabilities of 1 for all individuals with an event and of 0 for all individuals without an event. The Brier score has been criticized for being difficult to interpret, and its depends on the observed prevalence. For a null model where the estimated probability for each individual equals the observed prevalence, the Brier score is 0.25 when the prevalence is 0.5, and 0.09 when the prevalence is 0.01. The scaled Brier score, also known as the Brier Skill Score or the Index of Prediction Accuracy (IPA), is defined as 1 minus the Brier score divided by the Brier score for a null model.[79,80] For the scaled Brier score, a value of 0 corresponds to a null model and a value of 1 to a perfect model. The variants of loglikelihood and Brier score are all strictly proper (Table 2).

For continuous outcomes, the R-squared measures estimate the proportion of explained variation in the outcome. Because the concept of explained variation does not translate simply from a continuous to a binary outcome, several proposed R-squared measures for binary outcomes are referred to as pseudo R-squared measures.[81-83] The McFadden and Cox-Snell R-squared are based on the loglikelihood for the prediction model versus the loglikelihood for a null model. The Nagelkerke R-squared is a standardized version of the Cox-Snell R-squared to address its dependence on the prevalence and thus put it on the 0-1 scale.[84] Note that the Brier score is closely related to R-squared measures. Specifically, the scaled Brier score equals the sums-of-squares R-squared variant.[81-83] The R-squared measures are strictly proper.

The discrimination slope is the difference between the average estimated probability for individuals with an event and individuals without an event.[19,85] It has also been described as an R-squared variant called the coefficient of discrimination, and as the 'probabilistic AUROC'.[85,86] Despite its names, the discrimination slope is affected by calibration.[20] A related measure is the mean absolute prediction error (MAPE), which is the mean absolute difference between estimated probability and outcome. The discrimination slope and MAPE are improper.[35,87]

Because Brier and R-squared variants (except discrimination slope) measure overall performance and are strictly proper, they are useful to compare different models on the same dataset. Nevertheless, the values for these measures have no intuitive interpretation.

Overall performance measures for the case study are presented in Table 3. The plot that supports overall performance measures shows the distribution of the estimated probabilities for events and non-events separately. Figure 3 shows violin plots for the ADNEX model, showing that patients with a benign tumor mostly had very low estimated probabilities of malignancy. Patients with a malignancy commonly had a moderate to high estimated probabilities, but the distribution is less peaked.

## 6. Threshold-dependent evaluation

### 6.1. Classification measures

Classification performance is perfect when all individuals with an event have a probability above the decision threshold and all individuals without an event have a probability below the threshold. All classification measures are based on the cross-tabulation of predictions (e.g., low risk vs high risk) and outcomes (event vs no event), also called a contingency table or confusion matrix. At the commonly recommended threshold of 10% in the context of our case study,[26] ADNEX classified 578 patients as high risk, of which 414 had a malignant tumor (true positive, TP) and 164 had a benign tumor (false positive, FP). The remaining 316 patients were classified as low risk, of which 296 had a benign tumor (true negative, TN) and 20 had a malignant tumor (false negative, FN).

We divide measures for classification performance into summary measures and descriptive partial measures (Table S3-S4 for details and formulas).[88] Common descriptive partial measures are sensitivity or recall (the proportion of individuals with an event that are classified as high risk), specificity (the proportion of individuals without an event that are classified as low risk), PPV or precision (the proportion of individuals with an event among those classified as high risk), and negative predictive value (NPV, the proportion of individuals without an event among those classified as low risk). It is important to clearly understand the difference between sensitivity and specificity on the one hand, and PPV and NPV on the other. Sensitivity and specificity give probabilities conditional on the observed outcome, which is unknown at prediction time. For example, sensitivity is about individuals that eventually turn out to have the event, and estimates the proportion of those that were initially classified as high risk by the model. PPV and NPV are more clinically relevant, since they condition on the classification. For example, PPV is about individuals classified as high risk by the model, and estimates the proportion of those that turn out to have the event. Whereas sensitivity and specificity condition on the future and look back in time, PPV and NPV condition on the model prediction and look forward in time.

The most basic summary measure for classification is classification accuracy, which is the proportion of individuals who are correctly classified (TP and TN). Alternatives are the balanced accuracy and Youden index: the former is the average of sensitivity and specificity, the latter equals sensitivity plus specificity minus 1.[89,90] Another alternative is the kappa statistic, which expresses classification accuracy as the proportional increase beyond chance accuracy.[91] Chance accuracy is calculated by assuming that the distributions of the outcome and classifications are fixed and independent. Finally, the diagnostic odds ratio (DOR) shows how the odds of the event differ between those classified as low risk versus those classified as high risk.

The stronger the class imbalance, the easier it is to have higher classification accuracy than that of a model by simply classifying everyone as low risk if prevalence is <0.5 (or vice versa). This has led to the use of two alternative measures: F1 and Matthew's correlation coefficient (MCC). F1 is the harmonic mean of PPV/precision and sensitivity/recall.[92] F1 is reminiscent of the AUPRC, and shares its downsides: (1) F1 ignores TN so it conflates classification with utility, (2) F1 has no easy interpretation, and (3) the absolute value of F1 changes by simply switching the outcome labels (1 becomes 0, 0 becomes 1).[93,94] The same holds for the more general $F_{beta}$.[94] MCC is the Pearson correlation between classifications and observed outcomes.[95] Like F1, its value is hard to interpret intuitively.

For a given decision threshold *t*, all classification measures are improper (Supplementary Appendix 1). This is because a decision threshold implies specific misclassification costs, but these are not used in classification measures. Some classification measures are semi-proper if a decision threshold equal to 0.5 (classification accuracy) or to the true prevalence (balanced accuracy, Youden, F1) is used (Supplementary Appendix 1), but these are rarely the clinically relevant thresholds. In Supplementary Appendix 2, we describe theoretical relations between the summary classification measures for different values of the prevalence, sensitivity, and specificity.[88] Using similar theoretical exercises, it has been suggested that MCC provides the most comprehensive quantification of classification performance.[93,96]

Plots that relate to classification performance include the ROC and precision recall curves, because they plot partial classification measures across all possible decision thresholds. A limitation of these plots is that the thresholds are not easily visible (Figure 1).[41] An alternative plot is a classification plot, which has the probability threshold on the x-axis and one or more classification measures on the y-axis (Figure S2).[41]

For the ADNEX model, a threshold of 10% had a classification accuracy of 0.79 (95% confidence interval 0.77-0.82), F1 score of 0.82 (0.79-0.84), and MCC of 0.63 (0.58-0.67) (Table 3).

### 6.2. Clinical utility

In line with classic medical decision-analytic theory,[97] clinical utility focuses on the quality of decisions based on model classifications that correspond to a clinically relevant threshold. To assess utility, misclassification costs are explicitly considered. The most commonly used measure for clinical utility in prediction studies in healthcare is net benefit (NB) (Table S3-S4).[24,98,99] NB describes the net proportion of true positives as the proportion of true positives minus the weighted proportion of false positives. The weight is the ratio of the 'harm' of a false positive to the 'benefit' of a true positive (which is the opposite of the harm of a false negative). Harm and benefit are used in a broad sense.[21,22] This ratio is linked to the decision threshold.[21,22,24] For example, a threshold of 0.1 implies that we accept to apply the intervention in up to 10 individuals per intervention in an individual with the event (i.e. true positive). Put differently, we accept up to 9 unnecessary interventions (i.e. false positives) per true positive, which means that the harm of a false positive is 9 times smaller than the benefit of a true positive. Therefore, the weight used to calculate NB is the odds of t, t/(1-t). NB can also be calculated for the reference strategies of classifying everyone as high risk ('treat all') or as low risk ('treat none'). A model is suggested to be clinically useful if its NB is higher than that of both reference strategies. Lower AUC and miscalibration tend to reduce NB.[100,101] Miscalibration at t can even make a model harmful, which means that the model's NB at t is lower than that of a reference strategy. The maximum value of NB equals the prevalence. The 'standardized NB' divides NB by the prevalence, and is closely related to a measure called 'relative utility'.[29,102] Data scientists or clinicians may find standardized NB convenient, because its maximum value is 1 which facilitates comparison across validation studies of the model in populations with varying prevalence. In contrast, decision scientists argue that the prevalence is an important aspect of clinical utility and interpretability is lost by standardizing NB. NB is plotted in a decision curve with a range of reasonable decision thresholds on the x-axis (Figure 4A-B for the case study).[24,98] NB, whether standardized or not, is semi-proper: the probability estimates below (c.q. above) the threshold can be anything as long as they are below (c.q. above) the threshold.

A related measure is the Expected Cost (EC), where cost is again defined broadly.[103,104] It is a weighted sum of the proportion of false negatives and the proportion of false positives. The weights are the costs of a false negative and a false positive, respectively. EC does not make the link between the relative costs and the decision threshold: given the specification of the costs, the threshold that minimizes EC is chosen. For example, when the cost of a false negative is considered to be 9 times higher than the cost of a false positive, NB is calculated based on a decision threshold of 0.1 and a weight of odds(0.1). Instead, weights of 9 and 1 (or 0.9 and 0.1) are used to find the threshold that minimizes EC. The resulting threshold tends to be affected by miscalibration.[104] EC is a semi-proper measure that is insensitive to rank-preserving transformations of the probabilities. Of note, the minimization of EC can be visualized in ROC space.[105-107] If we normalize the costs to sum to one, we can plot EC for a range of reasonable normalized costs of a false positive or false negative (Figure 4C). Literature on EC allows the user to calculate EC for various assumed prevalences at model deployment, which may deviate from the prevalence in the validation dataset.[104,105] Plotting EC for various prevalences at deployment in combination with various misclassification costs (i.e. weights) leads to more complicated plots called cost curves.[105]

While it may seem that NB ignores TN and FN, and that EC ignores TP and TN, this is not the case. For example, harm of a FP refers to the overtreatment of individuals without the event, hence comparing negative consequences of an FP with positive consequences of a TN. Likewise, benefit of a TP refers to positive consequences of a TP versus negative consequences of a FN.[21,104,108]

For the ADNEX case study, NB was better than the reference strategies for all reasonable decision thresholds between 0.05 and 0.40 as explained in Box 1 (Figure 4A). The cost curve gives the same impression, with expected cost being lower than that of reference strategies for nearly all x-values. If we accept to intervene in up to 10 patients per true positive, NB of the model at t=0.1 is 0.44, which is better than the reference strategies (Table 3). For these relative costs, EC is minimized to 0.35 at a decision threshold of 6%. Using the observed prevalence in this dataset, the adopted relative costs relate to a value of 0.89 on the x-axis and of 0.07 on the y-axis of the cost curve. Following decision theory, the key concern is to check whether the model has better utility than the reference strategies and, if relevant, than competing models. To interpret the magnitude of the difference in NB, the test tradeoff can be used. We refer the reader to other literature on this topic.[108,109]

## 7. Results after recalibration

We updated the ADNEX model by fitting a logistic regression model of the outcome on the linear predictor ('logistic recalibration').[57,110] This is similar to Platt scaling, a well-known method in machine learning to improve calibration of predictions.[111,112] Graphical displays of model performance for the recalibrated model are shown in Figures S3-7. The calibration plot is closer to the diagonal after recalibration in the validation dataset (Figure S4). Table 3 provides all performance measures for the ADNEX model before and after recalibration. All strictly proper measures improved after recalibration. Semi-proper measures improved or did not change. The improper summary measures for classification (except DOR) worsened remarkably. Some partial classification measures improved (sensitivity, NPV), while others worsened (specificity, PPV). The improper measures for overall performance improved. The worsening of most improper performance measures after recalibration illustrates the importance of the properness concept. Of note, while logistic recalibration is a simple method to improve the model, it cannot improve discrimination because it is a rank preserving method.

## 8. Discussion

We distinguished between five domains in which performance of predictive AI models for use in medical practice can be evaluated: discrimination, calibration, overall performance, classification, and clinical utility. We presented 32 classic and contemporary performance measures across the five domains. We discussed two characteristics that a performance measure should exhibit: properness and a clear focus on either statistical or decision-analytic performance. Seventeen measures satisfied both characteristics. Twelve measures violated the first characteristic, two violated the second characteristic, and one measure (F1) violated both characteristics. We warn against the use of the 15 measures that do not satisfy both characteristics (Table 4). Improper measures should be avoided in model assessment, because such measures may mislead researchers rather than clarify performance of a model. Measures that conflate statistical and decision-analytic performance without properly accounting for misclassification costs should also be avoided, as their interpretation becomes unclear.

We argue that performance assessment for predictive AI for use in medical practice should primarily focus on discrimination, calibration, and utility.[113] Discrimination and calibration help both the modeler and clinician understand how the model might be improved. Poor discrimination indicates that other predictors may be sought to improve the distinction between individuals with and without the event. Although discrimination measures are more commonly reported than calibration measures,[114-117] calibration is a crucial aspect of model validation. Calibration has been labeled the “Achilles heel of predictive analytics”, because miscalibration can compromise predictive AI, leading to systematic over- or undertreatment.[100] Miscalibration is often not just a problem of the model, but a sign that we need to improve our understanding of the context in which the model is validated and used. Overall measures are influenced by elements of discrimination and calibration, and therefore are often not as informative as separate evaluations of discrimination and calibration. Utility is of interest for the decision maker and the patient: is there support that this model leads to better clinical decisions on average?

In particular, the core set of performance measures and plots that should be reported are AUROC, a smoothed calibration plot, a clinical utility measure such as net benefit with decision curve analysis, and a plot with probability distributions for each outcome category (Table 4). When internally validating a predictive AI model, calibration may be less important because model development and internal validation target the exact same population. Calibration is more important for external validation, when models are evaluated in different contexts and populations. Constructing an internally validated (i.e. optimism-corrected) calibration plot would still be useful, although a limited assessment of calibration using O:E ratio and calibration slope can be sufficient. Generally, O:E ratio should be close to 1 for properly developed models. Calibration slope can give an indication of overfitting. Although improper, the combination of PPV and NPV and/or the combination of sensitivity and specificity may be reported descriptively if desired, but always as an addition to the core set. Reported measures and plots should be accompanied by confidence intervals where possible, with the exception of clinical utility measures, a topic of recent debate.[118,119]

Class imbalance has received a lot of attention for model development and performance assessment. This attention was disproportionate: class imbalance is not a significant an issue as often claimed. The extent to which an outcome is imbalanced is not mathematically proportional to the extent to which misclassification costs are imbalanced. Class imbalance is related to the target population (an epidemiological feature of the data), whereas misclassification costs relate to the context of making decisions supported by the model (a clinical characteristic). Misclassification costs are informed by the nature and specifics of the medical decision that a model is supposed to support (e.g. whether or not to operate, take a biopsy, or start medication).[24,98,120] We have clinical utility measures to look into this in the context of decision-making. For this reason, we do not agree with researchers who recommend to use AUPRC, pAUROC and/or F1 instead of AUROC.[43-48,51,121] Ignoring TN is an important flaw of AUPRC and F1 for predictive AI that is developed to support clinical decisions. We do not make claims regarding other situations in healthcare where true negatives are not well-defined, such as lesion detection.[9]

Three topics related to performance assessment deserve emphasis: sample size, performance heterogeneity, and reporting transparency.

1) Sample size is crucial to assess performance with sufficient precision. Early recommendations suggested at least 100 to 200 events (individuals in the smallest outcome category), based on the precision to estimate AUROC, calibration statistics, and calibration plots.[55,122,123] More specific sample size calculations are now available for regression-based models.[124,125] When comparing calibration between models, often more data are needed.[126]

2) There is debate on how to deal with performance heterogeneity: patient populations, measurement procedures and treatment strategies can vary considerably between locations, settings, or time periods.[127-130] Heterogeneity in performance across locations can be quantified by meta-analyzing external validation studies.[128,131,132] In any case, it should be avoided to directly compare models using performance assessments that were derived from different datasets reflecting different populations from different settings.[38]

3) Comprehensive reporting of predictive AI modeling studies with respect to the aims, methods, results, and interpretation is imperative. This can be done by adhering to the TRIPOD checklists, including the 2024 TRIPOD+AI update (www.tripod-statement.org).[70,133,134] To avoid performance hacking, more attention should be given to publishing protocols in advance, as well as to sharing of analysis code and data where reasonably possible.[135]

Limitations of our overview include that we focused on the performance measures for binary outcomes. We expect many principles to hold for other types of outcomes as well, such as nominal, ordinal, time-to-event, or competing risk outcomes.[64,104,136-139] We also did not address counterfactual prediction (prediction under hypothetical interventions), which has gained traction recently.[140,141] Several other topics could not be discussed in depth. First, despite providing a comprehensive overview, it is impossible to discuss all measures, and research on performance measures remains ongoing. For example, calibration is an active area of research focusing on for example strong calibration, quantifying the degree of miscalibration, and uncertainty.[67,68,74,142,143] Second, we did not directly discuss model comparisons, although head-to-head comparisons of competing models on the same external validation dataset is of particular importance.[144] A specific topic related to model comparison involves studying the incremental value of adding a new predictor to an existing model.[145] While competing models can be evaluated using the same core set measures and visualizations, proper overall measures become more interesting for tasks like model selection and model comparison. Dedicated measures for evaluating competing models are available. One example is the net reclassification improvement (NRI).[146] Although widely used, NRI is an invalid performance measure because it is improper.[147,148]

In conclusion, we argue that performance measures should be proper, and should clearly focus on either purely statistical or adequate decision-analytic evaluation. To evaluate predictive AI models for use in medical practice, the core set of performance measures include AUROC, calibration plot, a clinical utility measure such as net benefit with decision curve analysis, and a plot showing the distribution of risk estimates.

**Contributors (CRediT roles)**

Conceptualization: BVC. Investigation: all authors. Methodology: all authors. Resources: BVC, DT. Data curation: BVC, DT. Formal analysis: BVC, LB. Software: BVC, LB. Validation: BVC, LB. Visualization: BVC, LB. Writing – original draft: BVC, EWS. Writing – review and editing: all authors. All authors agreed with submission of the manuscript for publication.

**Data sharing and ethics approval**

The retrospective use of data from the International Ovarian Tumor Analysis (IOTA) consortium was approved by the Research Ethics Committee from the University Hospitals (UZ) Leuven, IOTA's leading ethics committee (S64709). The code developed and utilized in this study is freely available from https://github.com/benvancalster/PerfMeasuresOverview. The full dataset is registered in the KU Leuven Research Data Repository (RDR) at https://doi.org/10.48804/TXL95Z, but cannot be publicly shared due to privacy considerations. However, we do provide risk estimates and outcomes (0 vs 1) on the GitHub page.

**Conflicts of Interest**

All authors declare no competing interests for this study.

**Acknowledgments**

BVC was supported by Research Foundation – Flanders (FWO) project G097322N, Internal Funds KU Leuven project C24M/20/064, and Kom Op Tegen Kanker (Stand up to Cancer) projects TRANS-IOTA and BIOC. GSC was supported by Cancer Research UK (programme grant: C49297/A27294), Medical Research Council Better Methods Better Research (grant MR/V038168/1) and the Engineering and Physical Sciences Research Council grant for "Artificial intelligence innovation to accelerate health research" (EP/Y018516/1). GSC is a National Institute for Health and Care Research (NIHR) Senior Investigator. The views expressed in this article are those of the author(s) and not necessarily those of the NIHR, or the Department of Health and Social Care. AJV was supported in part by grant P30CA008748 from the National Cancer Institute. LW was supported by the ZoNMW VIDI grant 09150172310023. KFK was supported by NIH / National Institutes of Diabetes and Digestive and Kidney Diseases (NIDDK) grant R01 DK093770. LB was supported by FWO project G097322N. GV acknowledges funding via the PEPR SN SMATCH France 2030 ANR-22-PESN-0003. KS was supported by the National Institutes of Health (NIH) projects R42AI177108, R01DK133226, and R01EB030492. TH-B was supported by the National Center for Advancing Translational Sciences of the NIH under Award Number UL1TR003142 and the National Library Of Medicine of the NIH under Award Number R01LM013362. The content is solely the responsibility of the authors and does not necessarily represent the official views of the National Institutes of Health. DT was supported by FWO projects G049312N/G0B4716N/12F3114N and Internal Funds KU Leuven project C24/15/037, and by Kom Op Tegen Kanker projects TRANS-IOTA and BIOC. DT is a senior clinical investigator of FWO.

Table 1. Five performance domains.

| Domain | Target question and focus | Focus |
|---|---|---|
| Probability-based | | |
| Discrimination | Does the model estimate higher probabilities in individuals with an event compared to individuals without an event? | Relative |
| Calibration | Do estimated probabilities correspond to observed event proportions? | Absolute |
| Overall | How close are estimated probabilities from the model (between 0 and 1) to actual outcomes (0 or 1)? | General |
| Threshold-dependent | | |
| Classification | Are individuals classified correctly corresponding to their observed outcome? | Binary |
| Clinical utility | Do classifications lead to better clinical decisions? | Clinical |

Table 2. Overview of performance measures and the assessment of the two key characteristics.

| **Measure** | **Characteristic 1: Properness** | **Characteristic 2: Focus** | **Characteristics met?** |
|---|---|---|---|
| Discrimination | | | |
| AUROC / AUC / concordance (c) statistic | + | + | Yes |
| AUPRC / AP | + | – | No |
| Partial AUROC | + | – | No |
| Calibration | | | |
| O:E ratio | + | + | Yes |
| Calibration intercept | + | + | Yes |
| Calibration slope | + | + | Yes |
| ECI | ++ | + | Yes |
| ICI | ++ | + | Yes |
| ECE | ++ | + | Yes |
| Overall performance | | | |
| Loglikelihood | ++ | + | Yes |
| Logloss/cross-entropy | ++ | + | Yes |
| Brier score | ++ | + | Yes |
| Scaled Brier / Brier Skill Score / IPA | ++ | + | Yes |
| McFadden R-squared | ++ | + | Yes |
| Cox-Snell R-squared | ++ | + | Yes |
| Nagelkerke R-squared | ++ | + | Yes |
| Coefficient of discrimination / Discrimination slope | – | + | No |
| Mean absolute prediction error (MAPE) | – | + | No |
| Classification (summary measures) | | | |
| Classification accuracy at t[#] | – | + | No |
| Balanced accuracy at t[#] | – | + | No |
| Youden index at t[#] | – | + | No |
| Diagnostic odds ratio at t[#] | – | + | No |
| Kappa at t[#] | – | + | No |
| F1 at t[#] | – | – | No |
| MCC at t[#] | – | + | No |
| Classification (partial measures) | | | |
| Sensitivity at t[#] | – | + | No |
| Specificity at t[#] | – | + | No |
| Positive predictive value (PPV or precision) at t[#] | – | + | No |
| Negative predictive value (NPV) at t[#] | – | + | No |
| Clinical utility | | | |
| Net benefit | + | + | Yes |
| Standardized NB | + | + | Yes |
| Expected cost | + | + | Yes |

[#] When one is interested in the optimal value of a summary classification measure over all possible decision thresholds, these measures are at best semi-proper: any rank preserving transformations of the estimated probabilities lead to the same optimal value.

Properness: ++, strictly proper; +, semi-proper; –, improper.

Focus: +, measure focuses either on purely statistical or on decision-analytic evaluation by properly addressing misclassification costs ; –, confusing mix of statistical and decision-analytic performance evaluation.

Table 3. Performance measures for the ADNEX model before and after recalibration.

| Measure | Before | After |
|---|---|---|
| Discrimination | | |
| AUROC / AUC / concordance (c) statistic | 0.911 (0.894, 0.927) | 0.911 (0.894, 0.927) |
| AUPRC / AP | 0.895 (0.862, 0.921) | 0.895 (0.862, 0.921) |
| Partial AUROC (sensitivity ≥0.8) | 0.141 (0.130, 0.151) | 0.141 (0.130, 0.151) |
| Calibration | | |
| O:E ratio | 1.228 (1.171, 1.288) | 1.000 (0.955, 1.046) |
| Calibration intercept | 0.810 (0.619, 1.006) | 0.000 (-0.180, 0.184) |
| Calibration slope | 0.934 (0.833, 1.051) | 1.000 (0.892, 1.126) |
| ECI | 0.105 (0.063, 0.160) | 0.002 (0.001, 0.017) |
| ICI | 0.094 (0.074, 0.118) | 0.014 (0.009, 0.038) |
| ECE | 0.091 (0.072, 0.117) | 0.017 (0.019, 0.050) |
| Overall performance | | |
| Loglikelihood | -370 (-407, -334) | -337 (-368, -307) |
| Logloss/cross-entropy | 370 (334, 407) | 377 (307, 368) |
| Brier score | 0.133 (0.118, 0.147) | 0.118 (0.106, 0.131) |
| Scaled Brier / Brier Skill Score / IPA | 0.469 (0.412, 0.527) | 0.526 (0.475, 0.576) |
| McFadden R-squared | 0.403 (0.343, 0.461) | 0.456 (0.405, 0.504) |
| Cox-Snell R-squared | 0.427 (0.379, 0.471) | 0.469 (0.429, 0.502) |
| Nagelkerke R-squared | 0.570 (0.505, 0.629) | 0.625 (0.573, 0.670) |
| Coefficient of discrimination / Discrimination slope | 0.509 (0.478, 0.540) | 0.525 (0.495, 0.556) |
| Mean absolute prediction error (MAPE) | 0.243 (0.226, 0.260) | 0.237 (0.222, 0.252) |
| Classification (t=0.10) | | |
| Classification accuracy at t | 0.794 (0.768, 0.819) | 0.691 (0.661, 0.723) |
| Balanced accuracy at t | 0.799 (0.776, 0.822) | 0.700 (0.677, 0.724) |
| Youden index at t | 0.597 (0.551, 0.643) | 0.399 (0.353, 0.448) |
| Diagnostic odds ratio at t | 37.4 (24.6, 68.5) | 43.3 (23.6 to 119) |
| Kappa at t | 0.592 (0.544, 0.639) | 0.392 (0.346, 0.442) |
| F1 at t | 0.818 (0.792, 0.843) | 0.756 (0.727, 0.782) |
| MCC at t | 0.625 (0.581, 0.667) | 0.480 (0.438, 0.522) |
| Sensitivity / recall | 0.954 (0.934, 0.974) | 0.984 (0.972, 0.993) |
| Specificity | 0.643 (0.603, 0.686) | 0.415 (0.370, 0.463) |
| Positive predictive value / precision | 0.716 (0.679, 0.753) | 0.614 (0.577, 0.650) |
| Negative predictive value | 0.937 (0.911, 0.964) | 0.965 (0.938, 0.986) |
| Clinical utility[a] | | |
| Net benefit (t=0.10) | 0.443 (0.411, 0.475) | 0.444 (0.411, 0.478) |
| Standardized net benefit (t=0.10) | 0.912 (0.892, 0.932) | 0.915 (0.900, 0.930) |
| Expected cost (costs 9:1)[b] | 0.355 (0.274, 0.376) | 0.355 (0.274, 0.376) |

[a] For clinical utility in particular, the use of confidence intervals and p-values for measures of clinical utility contradicts the principles of decision analysis.[119]

[b] EC was minimized at a decision threshold of 0.06 for the original model and 0.15 for the recalibrated model.

Table 4. Guidance and final comments per measure, considering a prediction model to support clinical decision-making.

| MEASURE/PLOT | GUIDANCE | COMMENT |
|---|---|---|
| | DISCRIMINATION | |
| AUROC | Recommended | Quantifies discrimination, which is a key component of statistical model performance. |
| AUPRC, pAUROC | Inadvisable | These measures attempt to move beyond a statistical assessment, but violate decision-analytic principles. |
| ROC curve, PR curve | Not inadvisable, but not essential either | These plots provide limited additional information over AUROC. |
| | CALIBRATION | |
| O:E ratio | Not inadvisable, but not essential either | An interpretable measure, but only a partial assessment of calibration; for internal validation, O:E ratio is often (close to) 1. |
| Calibration intercept, calibration slope | Not inadvisable, but not essential either | These measures are harder to interpret and provide a partial assessment of calibration; for internal validation, quantifying calibration slope can be used as in indication of overfitting.[66] |
| ECI, ICI, ECE | Not inadvisable, but not essential either | These measures summarize the smoothed (or grouped in case of ECE) calibration plot, concealing the nature and direction of miscalibration. |
| Calibration plot/reliability diagram | Recommended | This is by far the most insightful approach to assess calibration, in particular when smoothing rather than grouping is used; for internal validation, a plot is preferred but merely reporting the calibration slope is acceptable; for external validation a calibration plot is strongly recommended, with indications of uncertainty, e.g. by 95% confidence intervals. |
| | OVERALL | |
| Loglikelihood, Brier, R2 measures | Not inadvisable, but not essential either (in this work's context) | These proper measures are fine, yet it makes sense to conduct a separate evaluation of discrimination and calibration. Such measures are more convenient when comparing models, which was not the key focus of this work. |
| Discrimination slope, MAPE | Inadvisable | These measures are improper, which means that incorrect models can have better values for these measures than the correct model. |
| Risk distribution plots | Recommended | Displaying the distribution of the risk estimates for each outcome category provides valuable insights into a model's behavior. |
| | CLASSIFICATION | |
| Classification accuracy, balanced accuracy, Youden index, DOR, kappa, F1, MCC | Inadvisable | These measures are improper at clinically relevant decision thresholds; in addition, some measures are hard to interpret. |
| Sensitivity/recall and specificity | Not essential, can be descriptive if both reported together | While improper on their own, they can be reported descriptively if reported together. However, largely theoretical measures as they condition on the outcome that is predicted. |
| PPV/precision and NPV | Not essential, can be descriptive if both reported together | While improper on their own, they can be reported descriptively if reported together. PPV and NPV are more practical measures because they condition on the classification. |
| Classification plot | Not inadvisable, but not essential either | Classification plots plot could be presented descriptively, showing either sensitivity and specificity or PPV and NPV by threshold. |
| | CLINICAL UTILITY | |
| NB or standardized NB (with a decision curve), EC (with a cost curve) | Recommended | Important measures to quantify to what extent better decisions are made. Decision curves of NB allow one to show potential clinical utility at various clinically relevant decision thresholds relative to default decisions (and competing models). |

AUROC, area under the receiver operating characteristic (ROC) curve; AUPRC, area under the precision-recall (PR) curve; pAUROC, partial AUROC; ; ECI, estimated calibration index; ICI, integrated calibration index; ECE, expected calibration error; R2, R-squared; MAPE, mean absolute prediction error; DOR, diagnostic odds ratio; MCC, Matthew's correlation coefficient; PPV, positive predictive value; NPV, negative predictive value; NB, net benefit; EC, expected cost.

Figure 1. Receiver operating characteristic curve (panel A), precision recall curve (panel B), and plot to visualize pAUROC for the ADNEX model. For pAUROC, we considered sensitivities <0.8 to be unacceptably low.

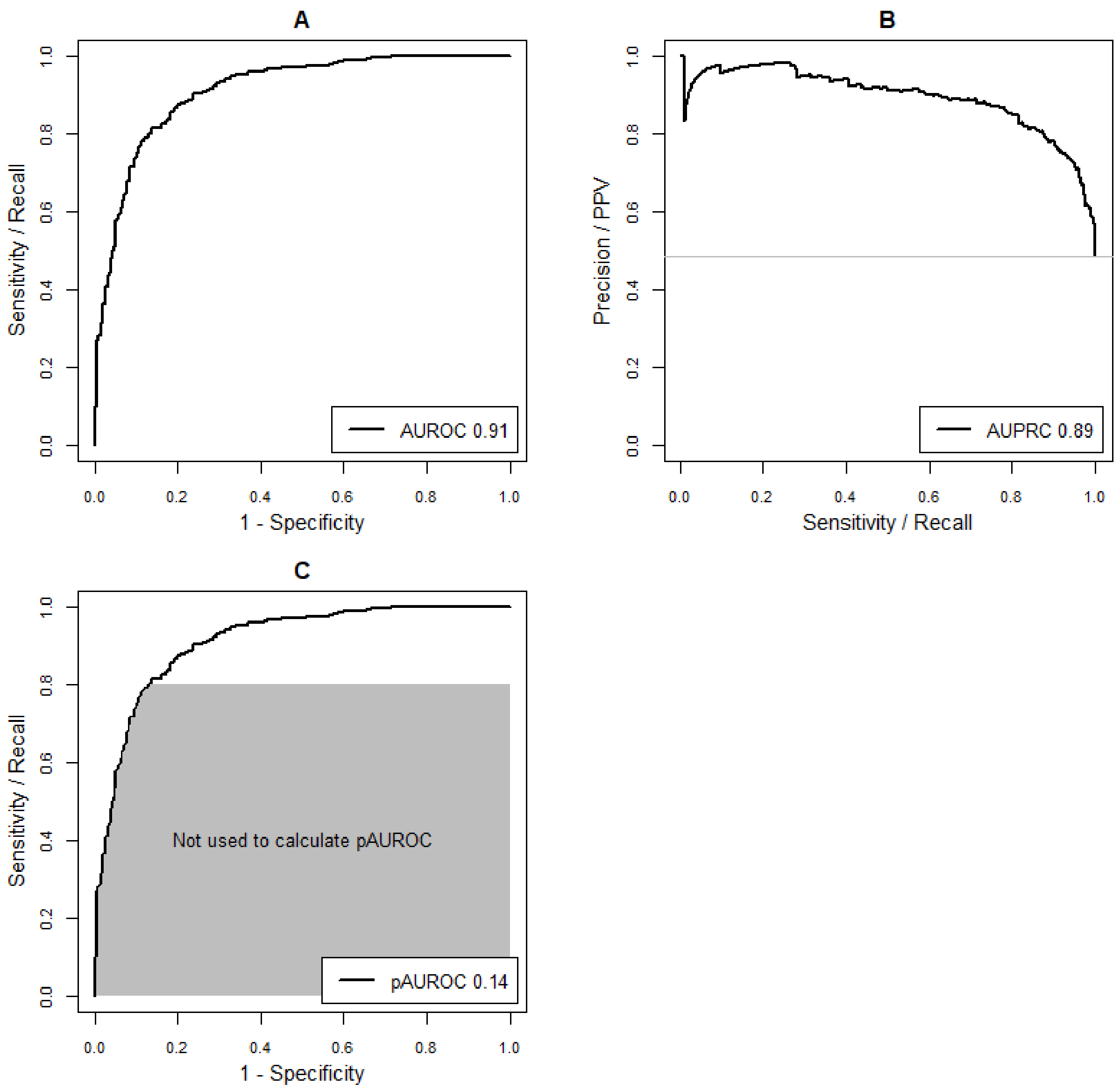

Figure 2. Calibration plot for the ADNEX model using 10 groups of equal sample size and using a loess smoother on the estimated probability.

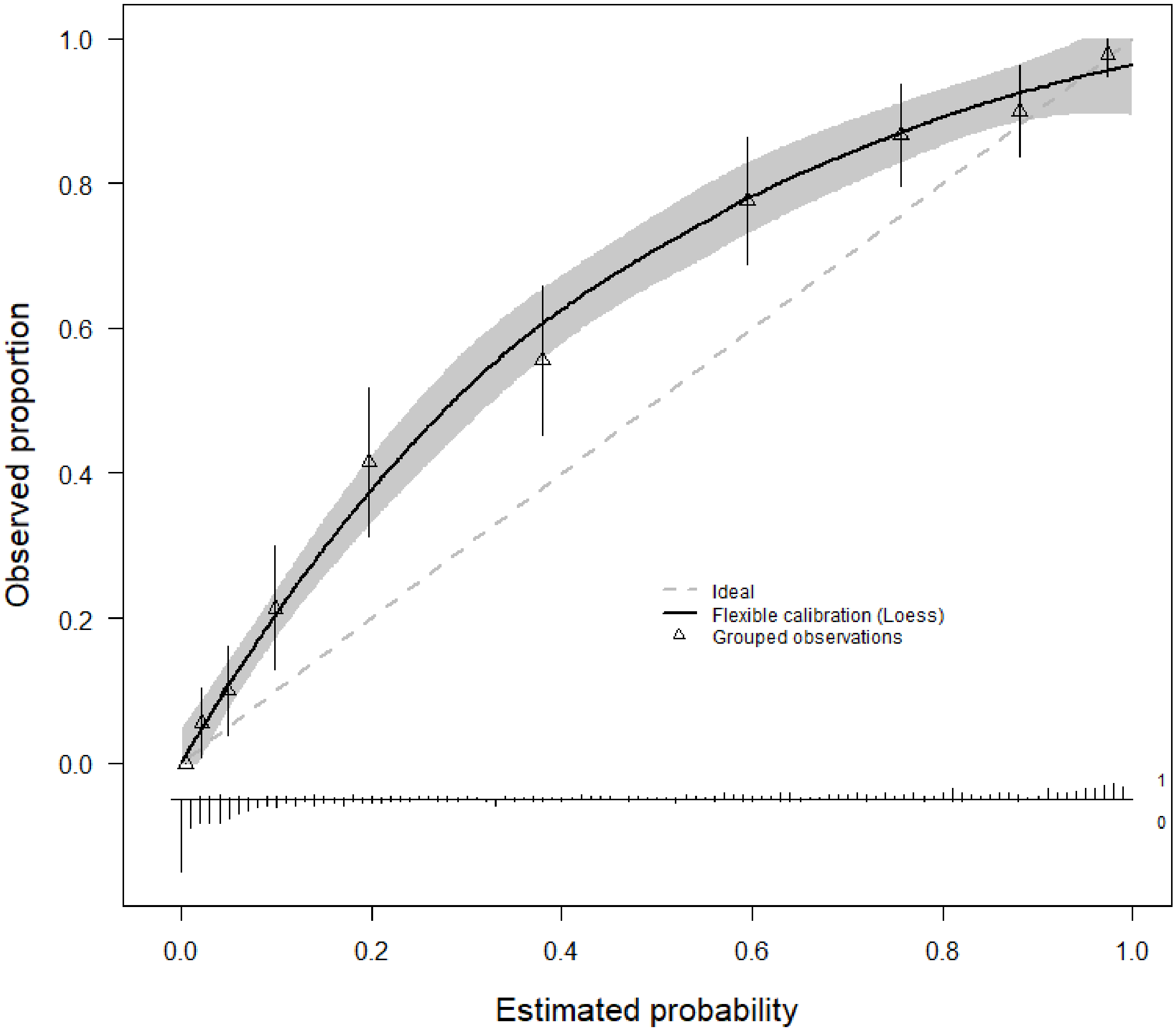

Figure 3. Violin and dot plots of the estimated probability of malignancy based on the ADNEX model.

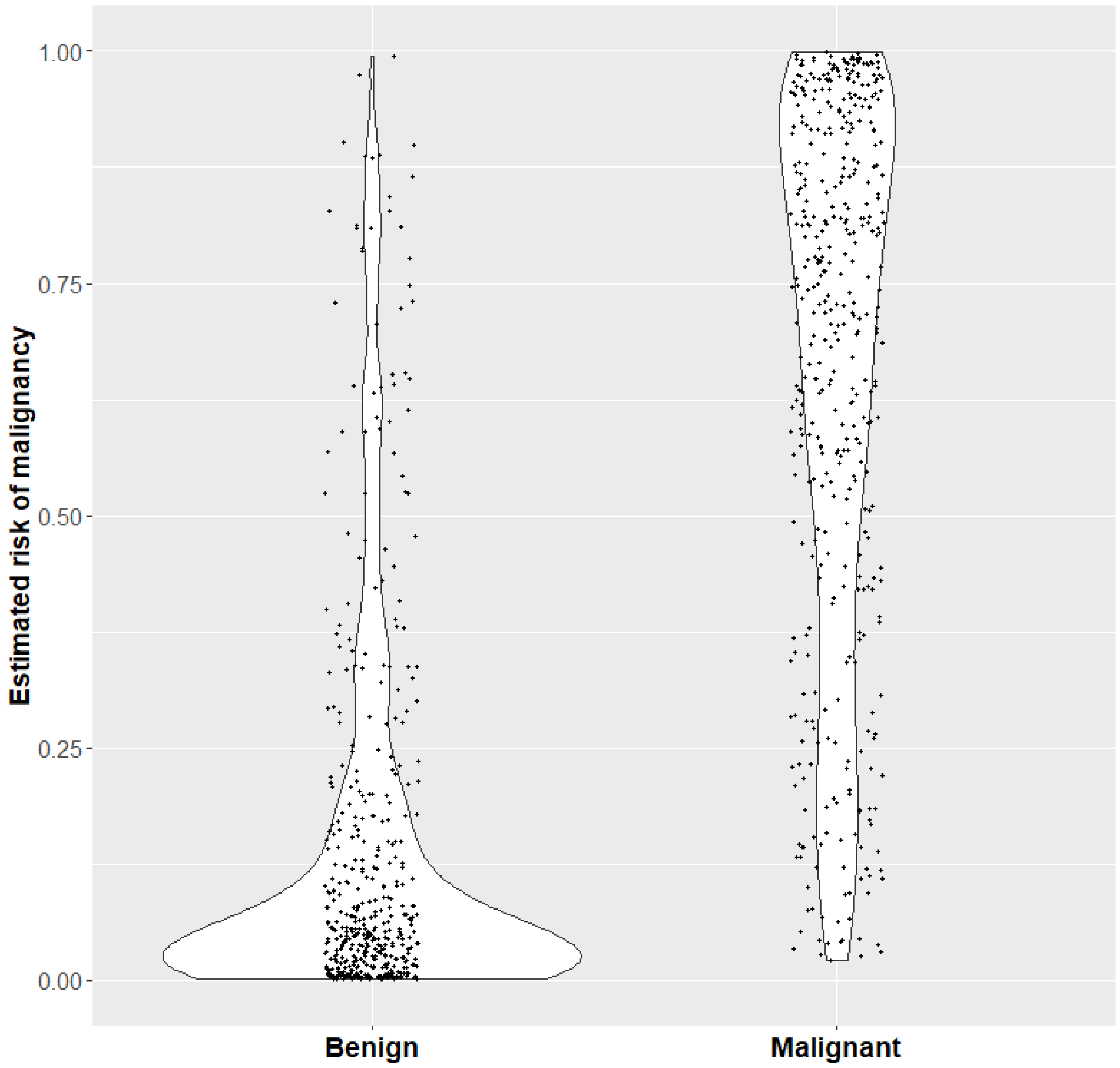

Figure 4. Decision curve with net benefit (A), standardized net benefit (B), and expected cost (C) for the case study. We show the full x-axis range for educational purposes. As explained in Box 1, a reasonable range of decision thresholds could be 0.05 to 0.40. This corresponds one on one with the normalized costs of a false negative on the curve for expected cost. When showing the decision curve in a validation study, the x-axis should be restricted to the reasonable range. Panel A also shows a smoothed curve using central moving averages. “All” (cq. “None”) refers to the net benefit or expect cost of the default strategy to classify all individuals as high (cq. low) risk.

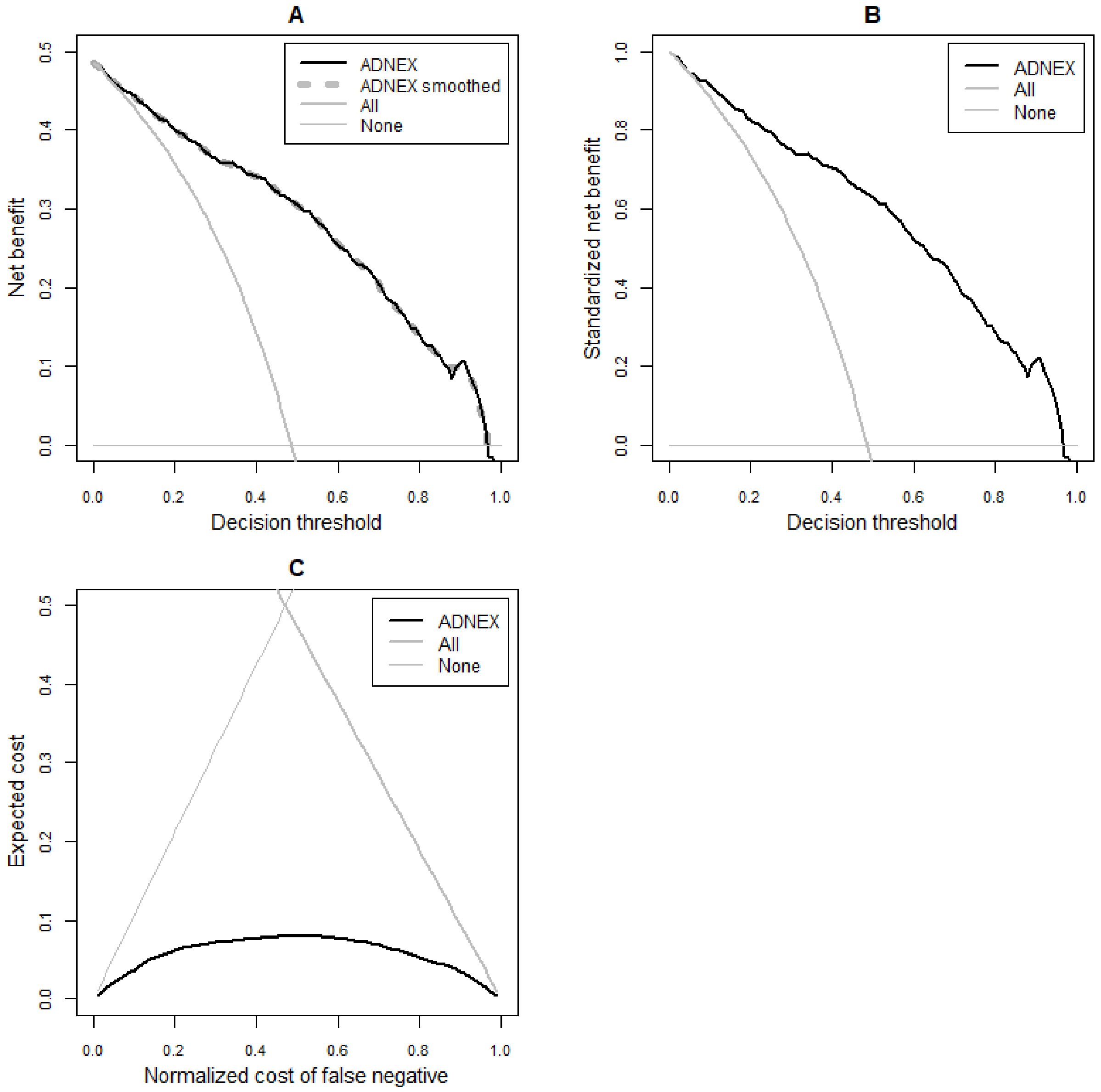

# Performance evaluation of predictive AI to support medical decisions: Overview and guidance

SUPPLEMENTARY MATERIAL

Ben VAN CALSTER[1,2,3], Gary S COLLINS[4], Andrew J VICKERS[5], Laure WYNANTS[1,2,6], Kathleen F KERR[7], Lasai BARREÑADA[1,3], Gael VAROQUAUX[8], Karandeep SINGH[9], Karel GM MOONS[10], Tina HERNANDEZ-BOUSSARD[11,12], Dirk TIMMERMAN[1,13], David J McLERNON[14], Maarten VAN SMEDEN[10], Ewout W STEYERBERG[3,10], for topic group 6 of the STRATOS initiative

1 Dept Development and Regeneration, KU Leuven, Leuven, Belgium

2 Leuven Unit for Health Technology Assessment Research (LUHTAR), KU Leuven, Leuven, Belgium

3 Dept Biomedical Data Sciences, Leiden University Medical Center, Leiden, the Netherlands

4 Centre for Statistics in Medicine, Nuffield Department of Orthopaedics, Rheumatology, and Musculoskeletal Sciences, University of Oxford, UK

5 Department of Epidemiology and Biostatistics, Memorial Sloan Kettering Cancer Center, New York, NY, USA

6 Department of Epidemiology, CAPHRI Care and Public Health Research Institute, Maastricht University, Maastricht, Netherlands

7 Department of Biostatistics, University of Washington School of Public Health, Seattle, WA, USA

8 Parietal project team, INRIA Saclay-Île de France, Palaiseau, France

9 Division of Biomedical Informatics, Department of Medicine, University of California, San Diego, CA, USA

10 Julius Centre for Health Sciences and Primary Care, University Medical Centre Utrecht, Utrecht University, Utrecht, Netherlands

11 Department of Medicine (Biomedical Informatics), Stanford University, Stanford, CA, USA

12 Department of Biomedical Data Science, Stanford University, Stanford, CA, USA

13 Department of Obstetrics and Gynecology, University Hospitals Leuven, Leuven, Belgium

14 Institute of Applied Health Sciences, University of Aberdeen, Aberdeen, UK

Table S1. Overview of common measures for discrimination, calibration, and overall performance.

| **Measure** | **Description** | **Target value**[a] | **Direction**[a] | **Comment** |
|---|---|---|---|---|
| DISCRIMINATION | | | | |
| AUROC / concordance (c) statistic | The probability that an individual with the event has a higher estimated probability than an individual without the event. | 1 | Higher | AUROC for model without predictive ability equals 0.5. |
| AUPRC / AP | Precision integrated over recall. | 1 | Higher | AUPRC for model without predictive ability equals the prevalence. |
| Partial AUROC | Area of part of the ROC curve with 'acceptable' sensitivity and/or specificity | <1 | Higher | Can be rescaled to 0.5 for a null model and 1 for a perfect model. |
| CALIBRATION | | | | |
| O:E ratio | Ratio of the number of individuals with an event and the expected number of individuals with an event according to the model. | 1 | Both | |
| Calibration intercept | Indication of whether probabilities are on average underestimated (intercept > 0), overestimated (<0), or correct (0). | 0 | Both | Adding the calibration intercept to the logit of the probabilities corrects for over- or underestimation. |
| Calibration slope | Indication of whether estimated probabilities are on average too extreme (slope>1), too modest (<1), or perfect (1). | 1 | Both | Multiplying the logit of the probabilities with the slope corrects the spread of the probabilities. |
| ECI | Mean squared difference between estimated probability and observed proportion from a calibration plot, divided by the value for a model without discriminatory ability. | 0 | Lower | Depends on smoother. This usually adds some noise such that ECI for a perfect model is not exactly 0. |
| ICI | Mean absolute difference between estimated probability and observed proportion from a calibration plot. | 0 | Lower | Depends on smoother. This usually adds some noise such that ICI for a perfect model is not exactly 0. |
| ECE | Weighted mean absolute difference between average probability and observed proportion from risk groups. | 0 | Lower | Depends on grouping. This usually adds some noise such that ECI for a perfect model is not exactly 0. |
| OVERALL | | | | |
| Loglikelihood | Loglikelihood of the outcome labels given the estimated probabilities. | 0 | Higher | |
| Logloss / Cross-entropy | Negative loglikelihood | 0 | Lower | Often average over N given. |
| Brier score | The average squared difference between the outcome labels and the estimated probabilities. | 0 | Lower | |
| Scaled Brier / Brier Skill Score / IPA | Brier score scaled to the value for a null model. | 1 | Higher | Asymptotically equivalent to Pearson R-squared. |
| McFadden R-squared | Proportion improvement in loglikelihood relative to a null model. | 1 | Higher | |
| Cox-Snell R-squared | More complex loglikelihood-based R-squared. | <1 | Higher | Target value depends on prevalence, which is inconvenient. |
| Nagelkerke R-squared | Cox-Snell R-squared scaled to the value for a perfect model. | 1 | Higher | |
| Discrimination slope | Difference in the mean event probability between individuals with an event and individuals without an event. | 1 | Higher | Despite the name, impacted by calibration. |
| MAPE | Mean absolute difference between outcome labels and estimated probabilities. | 0 | Lower | |

[a] The target value is the best possible value. Direction indicates whether higher or lower values are better, with 'both' meaning that the target value is not the maximum or minimum value, and that values above and below the target value are worse.

Table S2. Formulas for measures of overall, discrimination, and calibration performance.

| Measure | Definition/formula |
|---|---|
| OVERALL | |
| Loglikelihood ($l_M$) | $\sum_{n=1}^{N}[y_n * log(p_n) + (1-y_n) * log(1-p_n)]$ |
| Logloss / Cross-entropy | $-\sum_{n=1}^{N}[y_n * log(p_n) + (1-y_n) * log(1-p_n)]$ |
| Brier score | $N^{-1}\sum_{n=1}^{N}(p_n - y_n)^2$ |
| Scaled Brier score / Brier Skill score / IPA | $1-((N^{-1}\sum_{n=1}^{N}(p_n - y_n)^2)/\sum_{n=1}^{N}(\overline{y} - y_n)^2)$ |
| McFadden R-squared | $1-(l_M/l_0)$, with $l_0 = \sum_{n=1}^{N}[y_n * log(\overline{y}) + (1-y_n) * log(1-\overline{y})]$ |
| Cox-Snell R-squared | $1 - exp\left(N^{-1}2(l_0 - l_M)\right)$ |
| Nagelkerke R-squared | $(1 - exp\left(N^{-1}2(l_0 - l_M)\right))/(1 - exp(N^{-1}2l_0))$ |
| Discrimination Slope | $(\sum_{n=1}^{N} p_n y_n/\sum_{n=1}^{N} y_n) - (\sum_{n=1}^{N} p_n(1-y_n)/\sum_{n=1}^{N}(1-y_n))$ |
| MAPE | $N^{-1}\sum_{n=1}^{N}\lvert p_n - y_n\rvert = 1 - N^{-1}(\sum_{n=1}^{N} p_n y_n - \sum_{n=1}^{N} p_n(1-y_n))$ |
| DISCRIMINATION | |
| AUROC / Concordance (c) statistic | $(N_+N_-)^{-1}\sum_{n_+=1}^{N_+}\sum_{n_-=1}^{N_-} I\left(p_{n_+} > p_{n_-}\right)$ |
| Average precision (≈AUPRC) | $\sum_{n=1}^{N}\left(Recall_{t=p_n} - Recall_{t=p_{n+1}}\right)Precision_{t=p_n}$ |
| Partial AUROC | $(N_+N_-)^{-1}\sum_{n_+=1}^{N_+}\sum_{n_-=1}^{N_-} I\left(p_{n_+} > p_{n_-},\ p_{n_-} \in (q_1^-, q_0^-)\right)$ for a limited FPR range<br>$(N_+N_-)^{-1}\sum_{n_+=1}^{N_+}\sum_{n_-=1}^{N_-} I\left(p_{n_+} > p_{n_-},\ p_{n_+} \in (q_1^+, q_0^+)\right)$ for a limited sensitivity range |
| CALIBRATION | |
| O:E ratio | $\sum_{n=1}^{N} y_n/\sum_{n=1}^{N} p_n$ |
| Calibration intercept | $logit(Y) = \alpha + offset(L)$ |
| Calibration slope | $logit(Y) = \alpha + \beta L$ |
| Estimated Calibration Index | $N^{-1}\sum_{n=1}^{N}(p_n - o_n)^2/\sum_{n=1}^{N}(p_n - \overline{y})^2$ |
| Integrated Calibration Index | $N^{-1}\sum_{n=1}^{N}\lvert p_n - o_n\rvert$ |
| Expected Calibration Error | $\sum_{g=1}^{G}\frac{N_g}{N}\left\lvert\overline{p}_g - \overline{y}_g\right\rvert$ |

The dataset has $N$ individuals, $n = 1, \ldots, N$. $y_n$ is the outcome label (1 for event, 0 for no event) and $p_n$ the estimated probability of the event for individual n. $L$ is the linear predictor (i.e. logit of probability estimate). There are $N_+$ individuals with the event ($n_+ = 1, \ldots, N_+$) and $N_-$ individuals without the event ($n_- = 1, \ldots, N_-$). $p_{n_+}$ is the estimated probability of the event for individual $n_+$ with the event, $p_{n_-}$ is the estimated probability of the event for individual $n_-$ without the event. $Recall_{t=p_n}$ is recall/sensitivity when the threshold equals $p_n$, $Precision_{t=p_n}$ is precision when the threshold equals $p_n$. $Recall_{t=p_{n+1}}$ is 0 by definition. $q_1^-$ is the probability of event that gives the highest acceptable FPR, $q_0^-$ the probability of event that gives the lowest acceptable FPR, $q_1^+$ the probability of event that gives the highest acceptable sensitivity, $q_1^+$ the probability of event that gives the lowest acceptable sensitivity.

FPR, false positive rate (1 minus specificity);

Table S3. Overview of common measures for classification and clinical utility.

| Measure | Description | Target value[a] | Direction[a] | Comment |
|---|---|---|---|---|
| CLASSIFICATION (SUMMARY MEASURES) | | | | |
| Classification accuracy | The proportion of correctly classified individuals. | 1 | Higher | Depends on prevalence. |
| Balanced accuracy | The mean of sensitivity and specificity. | 1 | Higher | |
| Youden index | Sensitivity + specificity – 1. This is the difference between sensitivity and false positive rate. | 1 | Higher | |
| Diagnostic odds ratio | The ratio of true positive rate to false positive rate divided by the ratio of false negative rate to true negative rate. | Infinity | Higher | |
| Kappa | The relative proportion of correctly classified individuals, i.e. classification accuracy corrected for accuracy obtained by chance. | 1 | Higher | Depends on prevalence. |
| F1 | The harmonic mean of precision and recall. | 1 | Higher | Depends on prevalence. Ignores true negatives. |
| MCC | The phi correlation between the model's classifications and the actual outcomes. | 1 | Higher | Depends on prevalence. |
| CLASSIFICATION (PARTIAL MEASURES) | | | | |
| Sensitivity/recall | The proportion of individuals with an event that are correctly classified. | 1 | Higher | |
| Specificity | The proportion of individuals without event that are correctly classified. | 1 | Higher | |
| PPV/precision | The proportion of high risk individuals that have an event. | 1 | Higher | |
| NPV | The proportion of low risk individuals that do not have an event. | 1 | Higher | |
| CLINICAL UTILITY | | | | |
| Net Benefit | The net proportion of true positives. This is equivalent to the proportion of true positives in the absence of false positives. | Prevalence | Higher | Misclassification costs are explicitly linked to the decision threshold. |
| Standardized Net Benefit | The net sensitivity. This is equivalent to the sensitivity in the absence of false positives. | 1 | Higher | Misclassification costs are explicitly linked to the decision threshold. |
| Expected cost | Sum of the cost of false positives and the cost of false negatives. | 0 | Lower | Misclassification costs and decision threshold are disconnected. |

[a] The target value is the best possible value. Direction indicates whether higher or lower values are better.

Table S4. Formulas for measures of classification and clinical utility.

| Measure | Definition/formula |
|---|---|
| **CLASSIFICATION (OVERALL)** | |
| Classification accuracy | $\frac{TP + TN}{N}$ |
| Balanced accuracy | $0.5 * Sensitivity + 0.5 * Specificity$ |
| Youden index | $Sensitivity + Specificity - 1$ |
| Diagnostic odds ratio | $\frac{TP * TN}{FP * FN}$ |
| Kappa | $\frac{Accuracy - Accuracy_{chance}}{1 - Accuracy_{chance}} = \frac{N(TP + TN) - (TP + FP) * (TP + FN) - (FP + TN) * (FN + TN)}{N^2 - (TP + FP) * (TP + FN) - (FP + TN) * (FN + TN)}$ |
| F1 | $\frac{2 * PPV * Sensitivity}{PPV + Sensitivity} = \frac{2 * TP}{2 * TP + FP + FN}$ |
| Matthew's correlation coefficient (MCC) | $\frac{TP * TN - FP * FN}{\sqrt{(TP + FP) * (TP + FN) * (TN + FP) * (TN + FN)}}$ |
| CLASSIFICATION (PARTIAL) | |
| Sensitivity (recall) | $\frac{TP}{TP + FN}$ |
| Specificity | $\frac{TN}{TN + FP}$ |
| PPV (precision) | $\frac{TP}{TP + FP}$ |
| NPV | $\frac{TN}{TN + FN}$ |
| CLINICAL UTILITY | |
| Net Benefit | $\frac{TP}{N} - \frac{FP}{N}\left(\frac{t}{1 - t}\right)$ |
| Standardized Net Benefit | $\frac{TP}{TP + FN} - \frac{FP}{TP + FN}\left(\frac{t}{1 - t}\right)$ |
| Expected cost | $FNR * Prev * C(-\|+) + FPR * (1 - Prev) * C(+\|-) = \frac{FN}{N}C(-\|+) + \frac{FP}{N}C(+\|-)$ |

TP, number of true positives; TN, number of true negatives; N, sample size; FP, number of false positives; FN, number of false negatives; PPV, positive predicitve value; NPV, negative predictive value; FNR, false negative rate; Prev, prevalence; FPR, false positive rate; $C(-|+)$, cost of a false negative; $C(+|-)$, cost of a false positive.

Figure S1. Calibration plots for ADNEX by menopausal status (premenopausal patients in panel A, postmenopausal patients in panel B).

**A** **B**

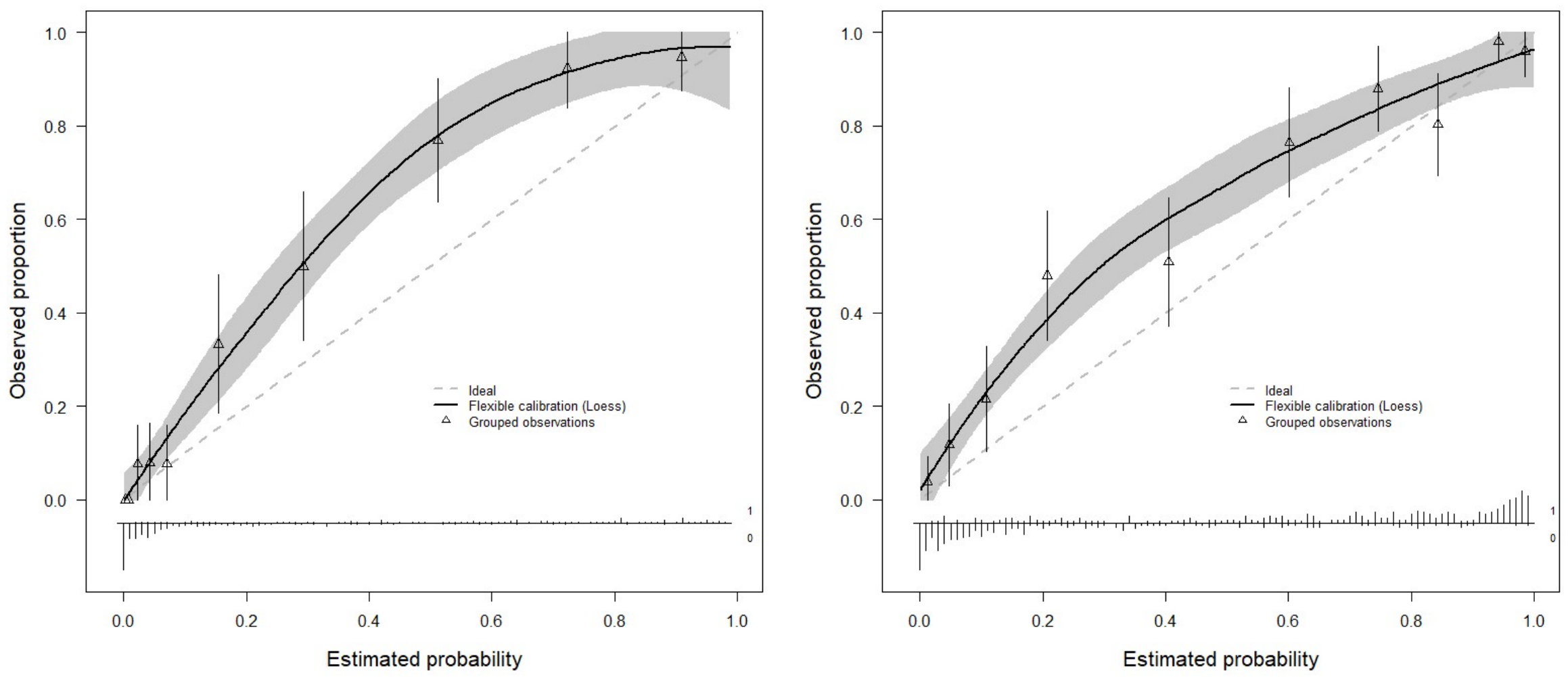

Figure S2. Classification plots for the ADNEX model showing sensitivity and 1 minus specificity (false positive rate) (panel A), PPV and 1 minus NPV (panel B), or PPV and Sensitivity (panel C).

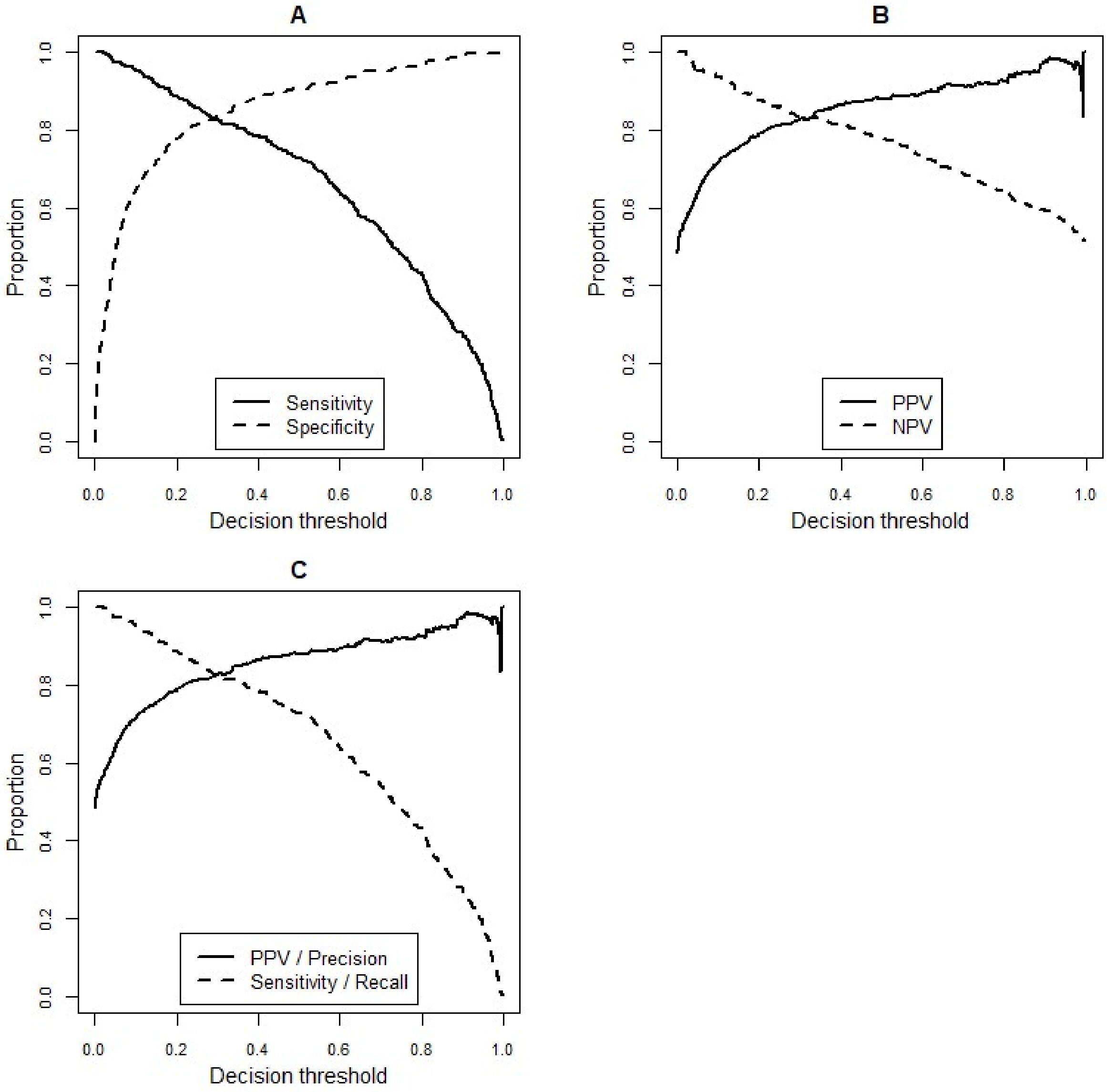

Figure S3. ROC curve (A), precision-recall curve (B) and plot for pAUROC (C) for the ADNEX model after recalibration.

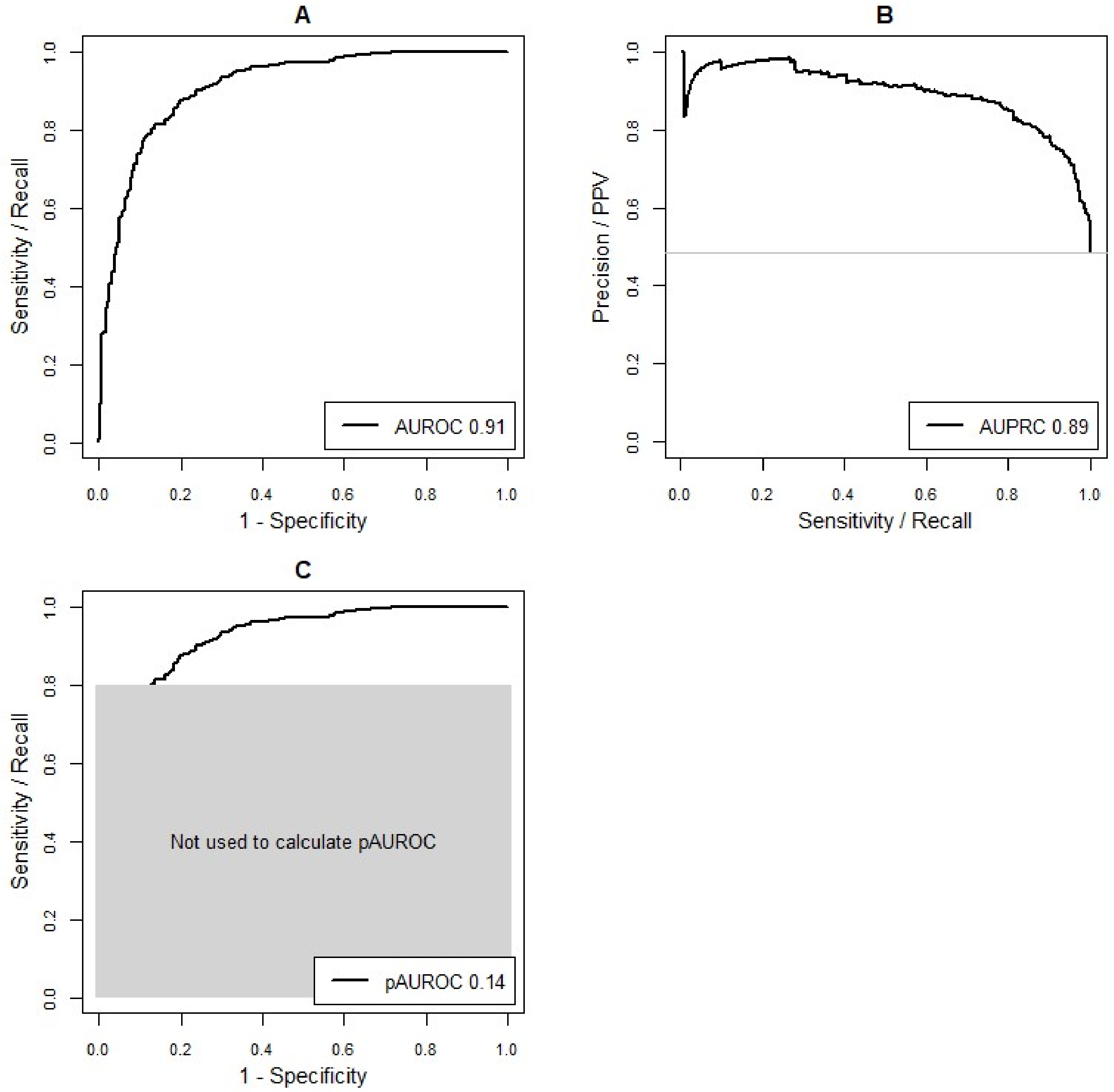

Figure S4. Calibration plot for the recalibrated ADNEX model using 10 groups of equal sample size and using a loess smoother on the estimated probability.

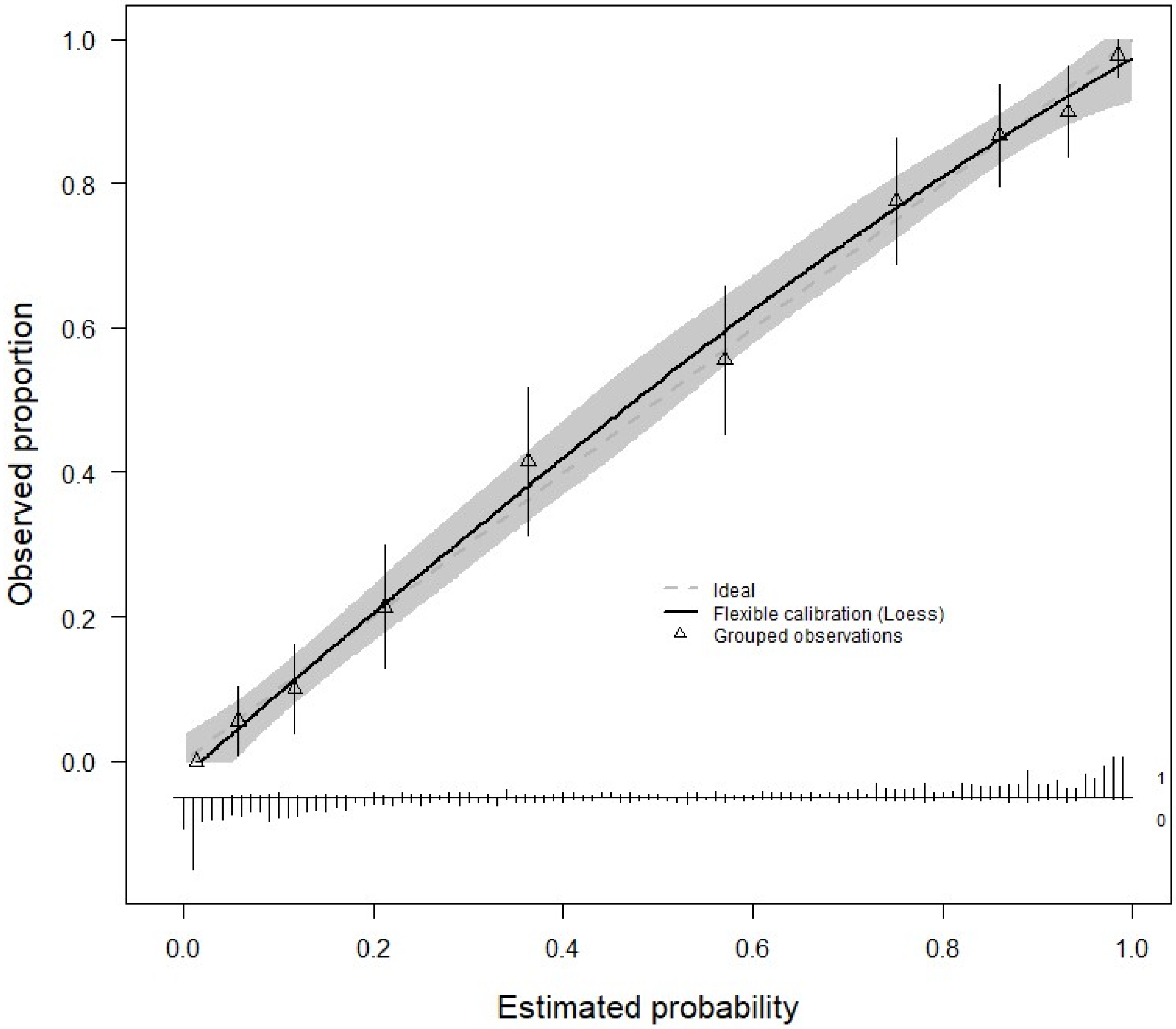

Figure S5. Violin and dot plots of the estimated probability of malignancy for the recalibrated model.

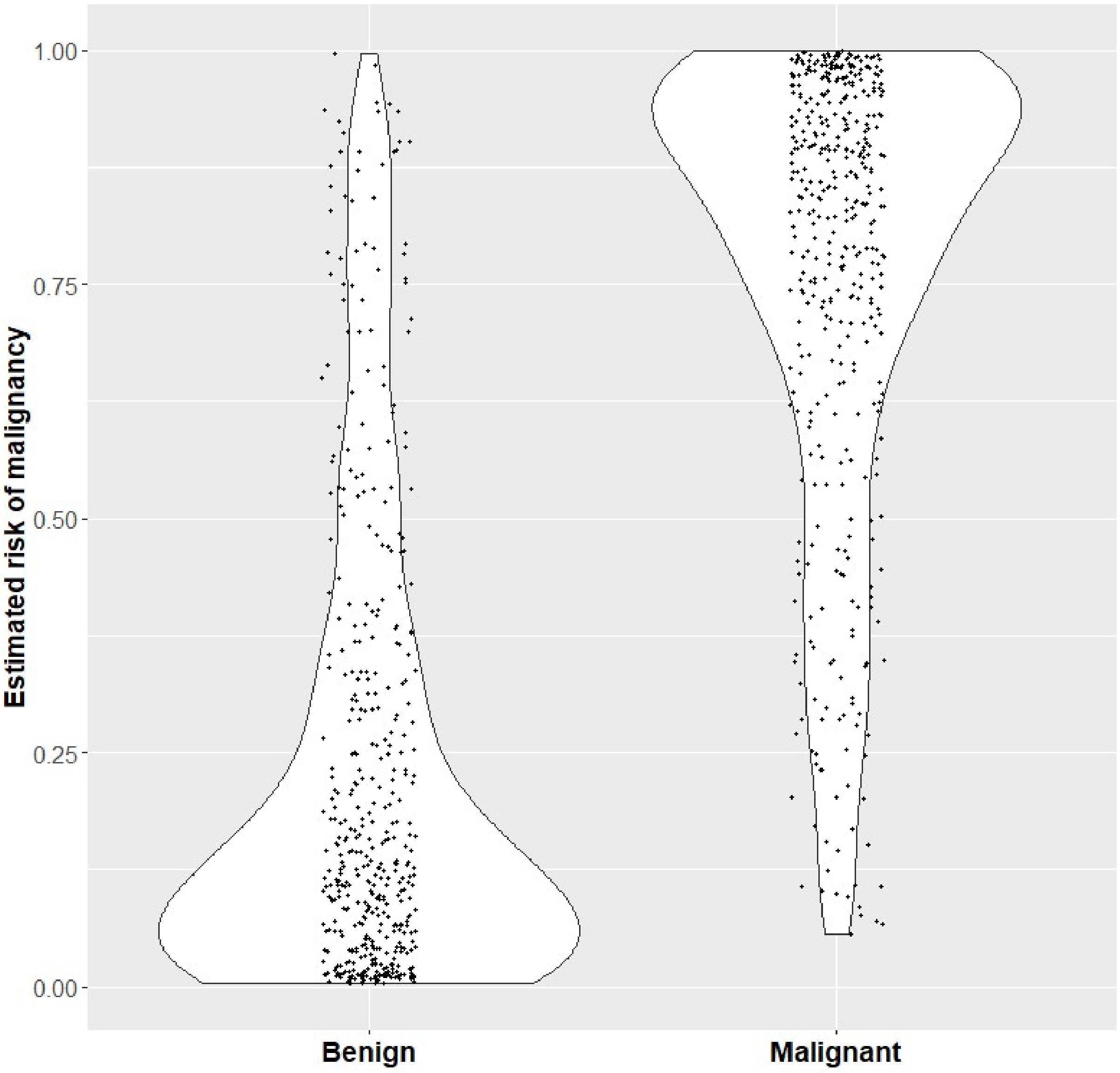

Figure S6. Classification plots for the recalibrated ADNEX model.

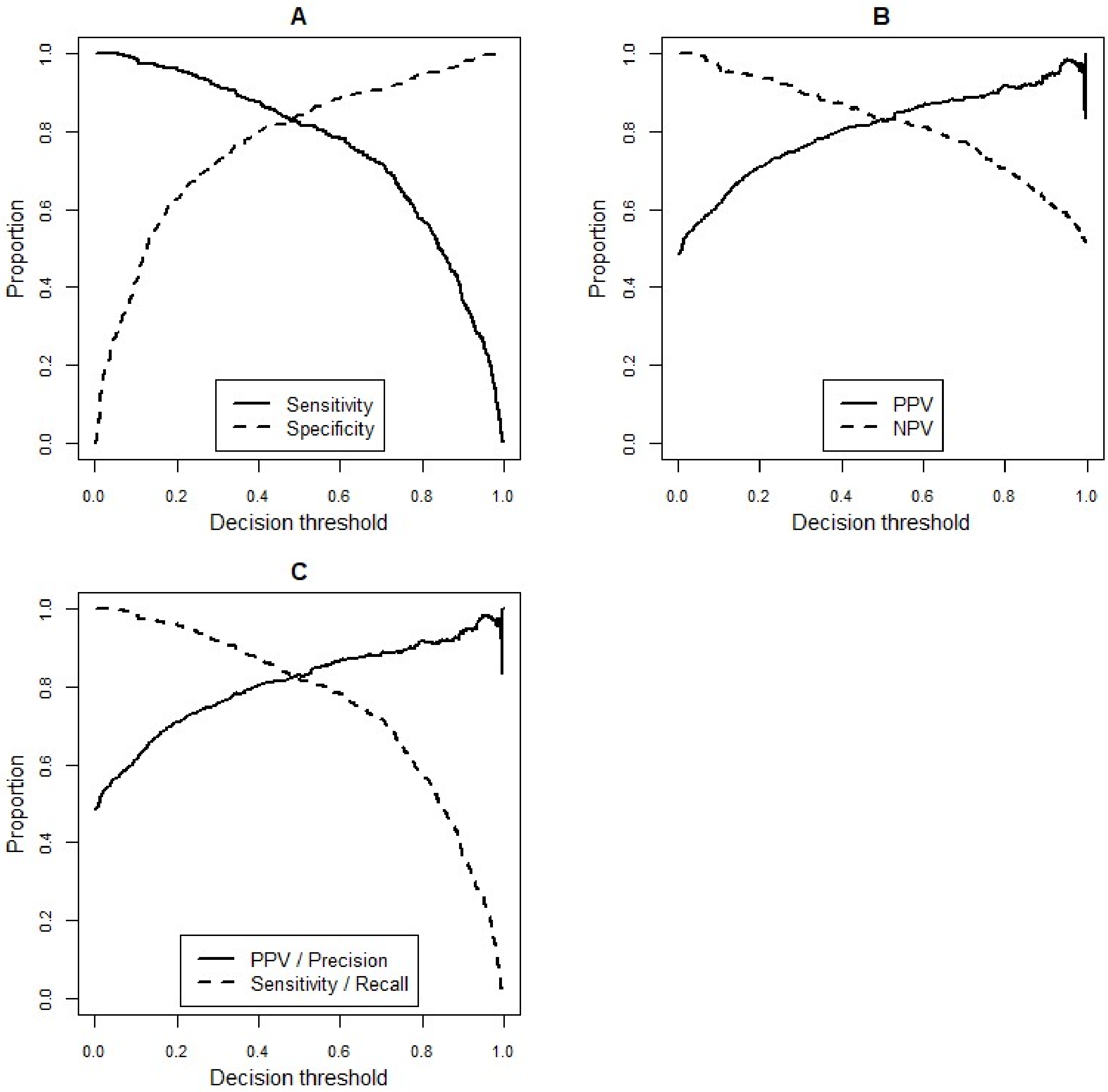

Figure S7. Decision curve using net benefit (A), standardized net benefit (B), and expected cost (C) for the recalibrated ADNEX model.

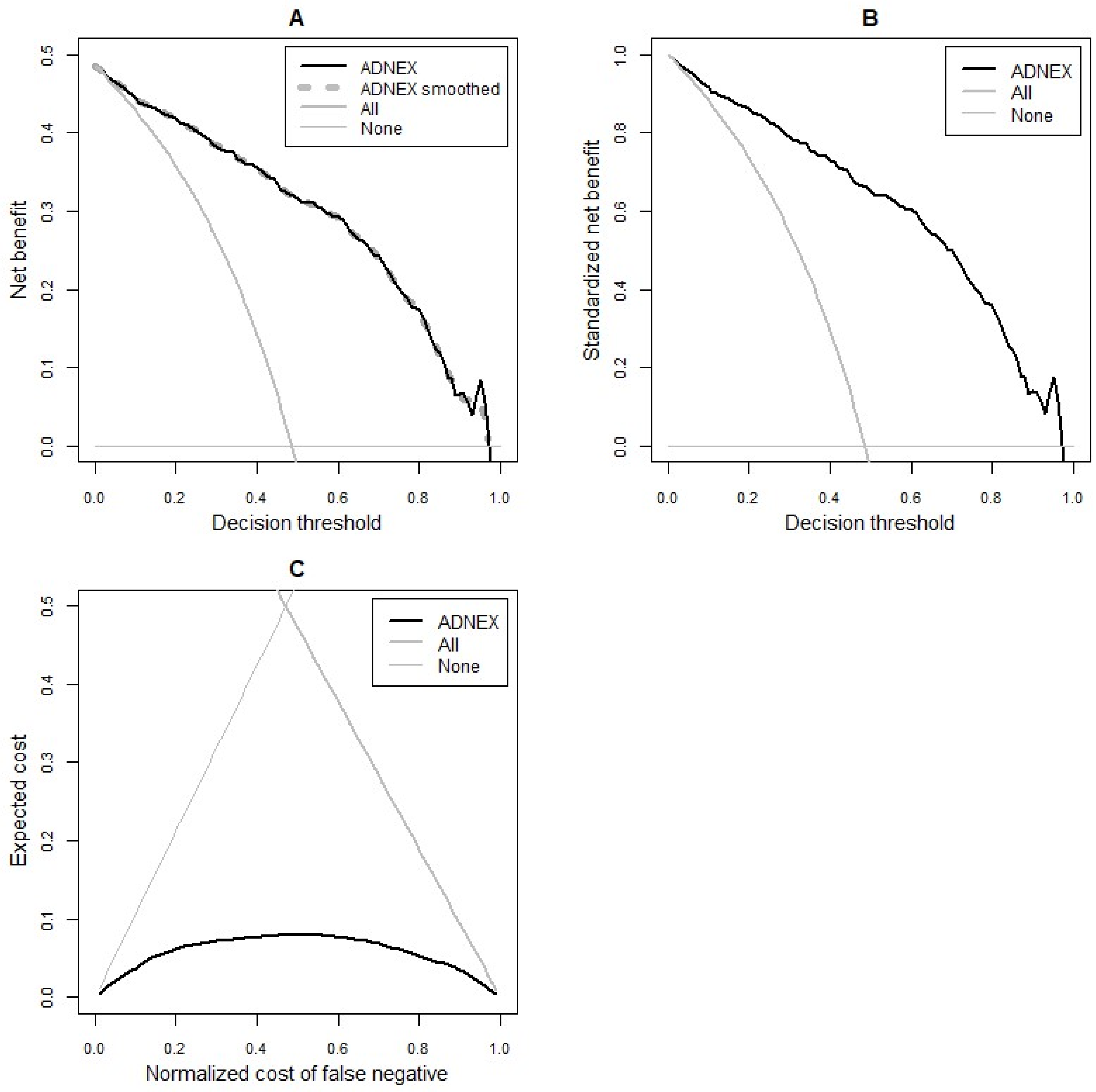

**Supplementary Appendix 1. Illustration of properness of the measures in this study.**

To illustrate the key property of properness, we consider a hypothetical situation with four continuous predictors that have a standard normal distribution and that are correlated with each other (Pearson correlation 0.4). The true model for outcome Y is a logistic model with intercept -1 and predictor coefficients of 0.74, 0.18, 0.18, and 0.18. This setup corresponds to a true AUROC of 0.746 and a prevalence of 0.304. The clinically relevant decision threshold on the probability of the event is set to 0.1.

We generate 2000 datasets of 1000 individuals, and the outcome Y for an individual is determined through a Bernoulli trial using the true probability of the event based on the logistic model above. We assess the performance measures on each dataset for the following models (see Figures A1-A2 for violin plots):

1. The true model, i.e. a model that uses the correct probability estimates.
2. A model using probability estimates equal to expit(trueLP + 0.75), with trueLP the linear predictor (i.e. logit(probability)) based on the true model and expit the inverse logit transformation.
3. A model using probability estimates equal to expit(trueLP – 1).
4. A model using probability estimates equal to expit(trueLP/1.3).
5. A model using probability estimates equal to expit(trueLP*2).
6. A model using probability estimates equal to expit(trueLP*2 – 1).
7. A model where true probabilities <0.1 are shrunk by a factor 10 (0.1 * true probability) and where true probabilities ≥0.1 are blown up using the transformation 1 – (0.1 * (1 – true probability)). Note that 0.1 is the decision threshold used to assess classification performance and clinical utility.
8. A model like model 7 but where true probabilities below the true prevalence are shrunk and true probabilities ≥ the true prevalence are blown up.
9. A model like model 7 but where true probabilities <0.5 are shrunk and true probabilities ≥0.5 are blown up.
10. A model where 0.04 is added or subtracted at random (allocation ratio 1:1) to true probabilities in the [0.051, 0.949] interval. This does not change the expected value of the average probability.

11. A model where model coefficients are 0.74, 0.74, 0.18, and 0.18, so the coefficient for predictor 2 is wrong.

Figure A1. Violin plot for the true model in this illustration.

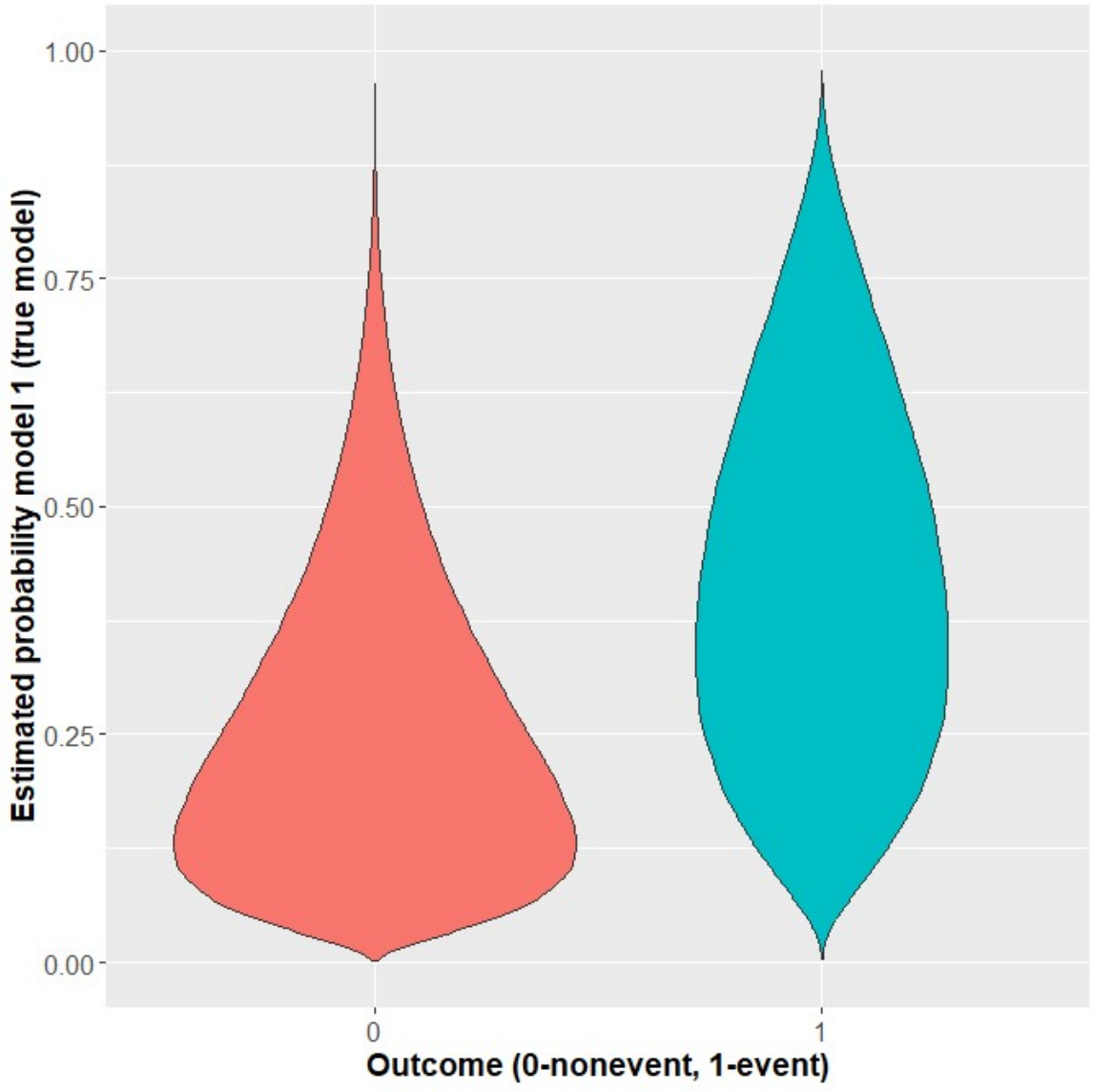

Figure A2. Violin plots for models 2 to 11 in this illustration.

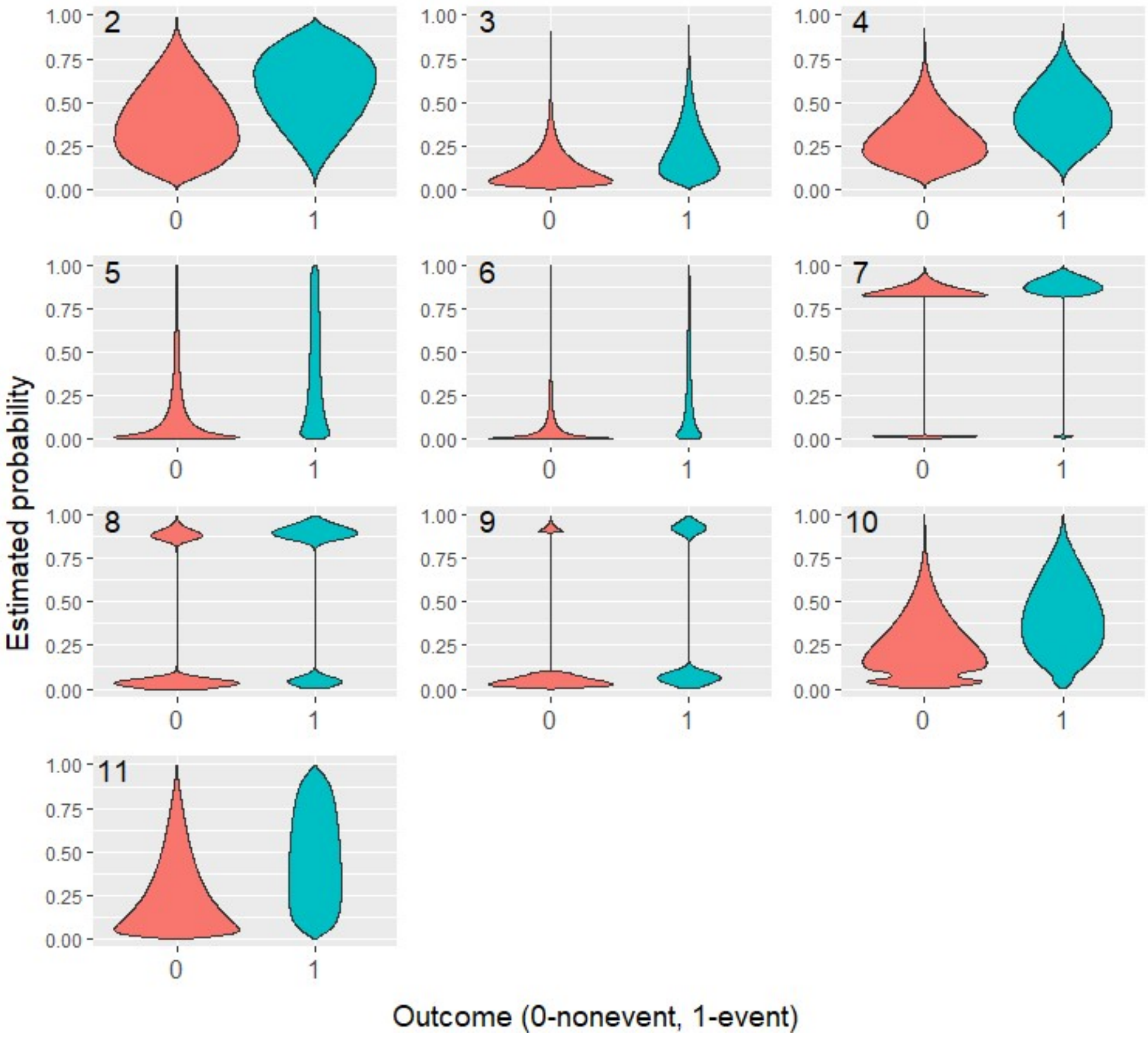

It is important that the variations of the true model do not depend in any way on the outcome. This would incur ‘leakage’. Take the extreme case as an example, in which you replace probabilities with the true outcome labels. This variation leads to perfect performance, way better than performance for the true model. This is caused by extreme leakage, which is not allowed to assess properness of a measure.

We report the average values of the performance measures over the 1000 datasets to approach their expected values (Table A1). This table shows that discrimination slope, MAPE, and the summary classification measures at a clinically relevant threshold (here 0.1) can give worse values for the true model than for other models. The discrimination measures, O:E ratio, calibration intercept and slope, and clinical utility measures can yield similar values for the true model and some other models. The other measures give the optimal value for the true model only.

Table A1. Average results across 2000 datasets of size 1000. Results that are better than those for the true model are highlighted in red (and the best result also in bold). When the result for the true model is the best, this value is highlighted in bold green and identical values for other models in green.

| Measure | 1 (true) | 2 | 3 | 4 | 5 | 6 | 7 | 8 | 9 | 10 | 11 |
|---|---|---|---|---|---|---|---|---|---|---|---|
| Loglikelihood | -530 | -583 | -609 | -537 | -605 | -732 | -1228 | -854 | -776 | -539 | -554 |
| Logloss | 0.530 | 0.583 | 0.609 | 0.537 | 0.605 | 0.732 | 1.228 | 0.854 | 0.776 | 0.539 | 0.554 |
| Brier | 0.177 | 0.200 | 0.203 | 0.179 | 0.190 | 0.212 | 0.442 | 0.270 | 0.232 | 0.180 | 0.185 |
| BSS | 0.162 | 0.055 | 0.039 | 0.153 | 0.102 | -0.004 | -1.095 | -0.276 | -0.098 | 0.151 | 0.123 |
| McFadden R2 | 0.137 | 0.050 | 0.008 | 0.126 | 0.015 | -0.192 | -1.002 | -0.391 | -0.265 | 0.122 | 0.098 |
| Cox-Snell R2 | 0.154 | 0.059 | 0.009 | 0.143 | 0.017 | -0.269 | -2.428 | -0.621 | -0.388 | 0.139 | 0.112 |
| Nagelkerke R2 | 0.218 | 0.084 | 0.014 | 0.203 | 0.024 | -0.379 | -3.437 | -0.879 | -0.548 | 0.197 | 0.160 |
| Discrimination slope | 0.163 | 0.182 | 0.110 | 0.138 | 0.216 | 0.155 | 0.140 | **0.320** | 0.231 | 0.163 | 0.200 |
| MAPE | 0.354 | 0.403 | 0.319 | 0.377 | 0.302 | **0.291** | 0.544 | 0.328 | 0.282 | 0.354 | 0.347 |
| AUROC | 0.746 | 0.746 | 0.746 | 0.746 | 0.746 | 0.746 | 0.746 | 0.746 | 0.746 | 0.737 | 0.733 |
| AUPRC | 0.566 | 0.566 | 0.566 | 0.566 | 0.566 | 0.566 | 0.566 | 0.566 | 0.566 | 0.556 | 0.548 |
| pAUC | 0.069 | 0.069 | 0.069 | 0.069 | 0.069 | 0.069 | 0.069 | 0.069 | 0.069 | 0.064 | 0.065 |
| O:E ratio | 1.000 | 0.678 | 1.941 | 0.904 | 1.338 | 2.300 | 0.398 | 0.748 | 1.588 | 1.000 | 0.934 |
| Calibration intercept | -0.001 | -0.751 | 0.999 | -0.167 | 0.631 | 1.631 | -2.577 | -1.357 | 1.439 | -0.001 | -0.136 |
| Calibration slope | 1.006 | 1.006 | 1.006 | 1.308 | 0.503 | 0.503 | 0.671 | 0.290 | 0.344 | 0.885 | 0.669 |
| ECI | 0.023 | 0.354 | 0.711 | 0.103 | 0.190 | 0.549 | 0.917 | 0.482 | 0.473 | 0.028 | 0.092 |
| ICI | 0.021 | 0.145 | 0.148 | 0.045 | 0.106 | 0.176 | 0.473 | 0.257 | 0.211 | 0.025 | 0.057 |
| ECE | 0.033 | 0.145 | 0.148 | 0.051 | 0.107 | 0.172 | 0.470 | 0.234 | 0.185 | 0.036 | 0.063 |
| Classification accuracy | 0.408 | 0.331 | 0.611 | 0.336 | 0.631 | 0.708 | 0.408 | 0.678 | **0.737** | 0.414 | 0.467 |
| Balanced accuracy | 0.567 | 0.518 | 0.667 | 0.522 | 0.673 | 0.675 | 0.567 | **0.679** | 0.624 | 0.568 | 0.599 |
| Youden index | 0.135 | 0.036 | 0.334 | 0.043 | 0.345 | 0.350 | 0.135 | **0.359** | 0.248 | 0.135 | 0.198 |
| Diagnostic odds ratio | 8.383 | 13.594 | 4.785 | **14.499** | 4.690 | 4.613 | 8.383 | 4.560 | 5.355 | 5.840 | 5.530 |
| Kappa | 0.088 | 0.022 | 0.267 | 0.027 | 0.284 | **0.336** | 0.088 | 0.321 | 0.285 | 0.089 | 0.136 |
| F1 | 0.500 | 0.475 | 0.559 | 0.477 | 0.562 | 0.551 | 0.500 | **0.563** | 0.437 | 0.499 | 0.517 |
| MCC | 0.190 | 0.099 | 0.311 | 0.109 | 0.319 | **0.338** | 0.190 | 0.333 | 0.308 | 0.183 | 0.227 |
| Net benefit (NB) | 0.231 | 0.229 | 0.210 | 0.229 | 0.204 | 0.161 | 0.231 | 0.183 | 0.096 | 0.229 | 0.228 |
| Standardized NB | 0.760 | 0.752 | 0.688 | 0.753 | 0.668 | 0.529 | 0.760 | 0.600 | 0.314 | 0.751 | 0.748 |
| Expected cost | 0.632 | 0.632 | 0.632 | 0.632 | 0.632 | 0.632 | 0.632 | 0.632 | 0.632 | 0.647 | 0.641 |

Some other observations are worth mentioning. First, models 6-9 have negative values for the scaled Brier score and the likelihood-based R-squared measures, suggesting that these models are worse than a null model. Still, these models have better performance than the true model for some improper measures. Second, the calibration intercept for the true model was not 0 up to the third decimal even after assessing it on 2 million individuals in total. This measure has more variability than the O:E ratio. Third, the moderate calibration measures (ECI, ICI, ECE) were not 0 for the true model. This is probably the consequence of the dependence on smoothing (ECI and ICI) or grouping (ECE) to obtain 'observed' proportions, adding noise to the individual datasets of size 1000. Fourth, the semi-proper discrimination measures yield exactly the same results for any model that preserves the rank-order of individuals based on the estimated probabilities. Only models 9 and 10 do not preserve the ranks, and this leads to worse values for semi-proper measures. Fifth, although classification measures are improper at a given clinically relevant decision threshold, some are semi-proper if this threshold happens to coincide with the true prevalence (balanced accuracy, Youden index, F1) or with 0.5 (classification accuracy) (Table A2). This will rarely be the case in real applications.

Table A2. Average results for classification performance across 2000 datasets of size 1000, when either the true prevalence (0.304) or 0.5 were used as the decision threshold. Results that are better than those for the true model are highlighted in red (and the best result also in bold). When the result for the true model is the best, this value is highlighted in bold green and identical values for other models in green.

| Measure | 1 (true) | 2 | 3 | 4 | 5 | 6 | 7 | 8 | 9 | 10 | 11 |
|---|---|---|---|---|---|---|---|---|---|---|---|
| Classification accuracy, t=0.304 | 0.678 | 0.529 | 0.735 | 0.636 | 0.724 | 0.736 | 0.408 | 0.678 | **0.737** | 0.672 | 0.663 |
| Balanced accuracy, t=0.304 | **0.679** | 0.633 | 0.604 | 0.674 | 0.664 | 0.614 | 0.567 | 0.679 | 0.624 | 0.674 | 0.670 |
| Youden index, t=0.304 | **0.359** | 0.267 | 0.208 | 0.348 | 0.328 | 0.228 | 0.135 | 0.359 | 0.248 | 0.348 | 0.340 |
| Diagnostic odds ratio, t=0.304 | 4.560 | 5.402 | 5.780 | 4.674 | 4.750 | 5.552 | **8.383** | 4.560 | 5.355 | 4.346 | 4.190 |
| Kappa, t=0.304 | 0.321 | 0.193 | 0.249 | 0.288 | **0.335** | 0.268 | 0.088 | 0.321 | 0.285 | 0.310 | 0.299 |
| F1, t=0.304 | **0.563** | 0.537 | 0.382 | 0.563 | 0.530 | 0.411 | 0.500 | 0.563 | 0.437 | 0.558 | 0.554 |
| MCC, t=0.304 | 0.333 | 0.272 | 0.290 | 0.321 | **0.336** | 0.299 | 0.190 | 0.333 | 0.308 | 0.323 | 0.314 |
| | | | | | | | | | | | |
| Classification accuracy, t=0.5 | **0.737** | 0.689 | 0.711 | 0.737 | 0.737 | 0.726 | 0.408 | 0.678 | 0.737 | 0.734 | 0.723 |
| Balanced accuracy, t=0.5 | 0.624 | 0.679 | 0.529 | 0.624 | 0.624 | 0.569 | 0.567 | **0.679** | 0.624 | 0.623 | 0.644 |
| Youden index, t=0.5 | 0.248 | 0.358 | 0.058 | 0.248 | 0.248 | 0.137 | 0.135 | **0.359** | 0.248 | 0.246 | 0.289 |
| Diagnostic odds ratio, t=0.5 | 5.355 | 4.561 | **12.380** | 5.355 | 5.355 | 7.045 | 8.383 | 4.560 | 5.355 | 5.106 | 4.427 |
| Kappa, t=0.5 | 0.285 | **0.328** | 0.079 | 0.285 | 0.285 | 0.175 | 0.088 | 0.321 | 0.285 | 0.282 | 0.306 |
| F1, t=0.5 | 0.437 | 0.561 | 0.121 | 0.437 | 0.437 | 0.270 | 0.500 | **0.563** | 0.437 | 0.437 | 0.493 |
| MCC, t=0.5 | 0.308 | **0.336** | 0.171 | 0.308 | 0.308 | 0.246 | 0.190 | 0.333 | 0.308 | 0.303 | 0.310 |

**Supplementary Appendix 2. Theoretical relations between summary measures for classification.**

We made all 9261 combinations of the following values for prevalence, sensitivity, and specificity: 0.01, 0.05 to 0.95 in steps of 0.05, and 0.99. Using these three quantities, the proportion true positives, false positives, true negatives, and false negatives can be calculated. For each combination, we calculated classification accuracy (Acc), balanced accuracy (BAR), Youden index, diagnostic odds ratio (DOR), kappa, F1, and Matthews correlation coefficient (MCC).

Figure A3 presents histograms, scatter plots and Spearman correlations of the measures across all combinations. BAR and Youden are by definition perfectly correlated. Spearman correlations between BAR/Youden, DOR, Kappa, and MCC varied between 0.93 and 0.98. These measures had correlations between 0.81 and 0.86 with classification accuracy. The reason for the lower correlations is that only classification accuracy makes no distinction between individuals with an event and individuals without. F1 correlated only between 0.64 and 0.69 with all other measures, because F1 ignores true negatives.

Figure A3. Histograms, scatter plots and Spearman correlations between summary measures for classification for 9261 theoretical combinations of prevalence, sensitivity, and specificity.

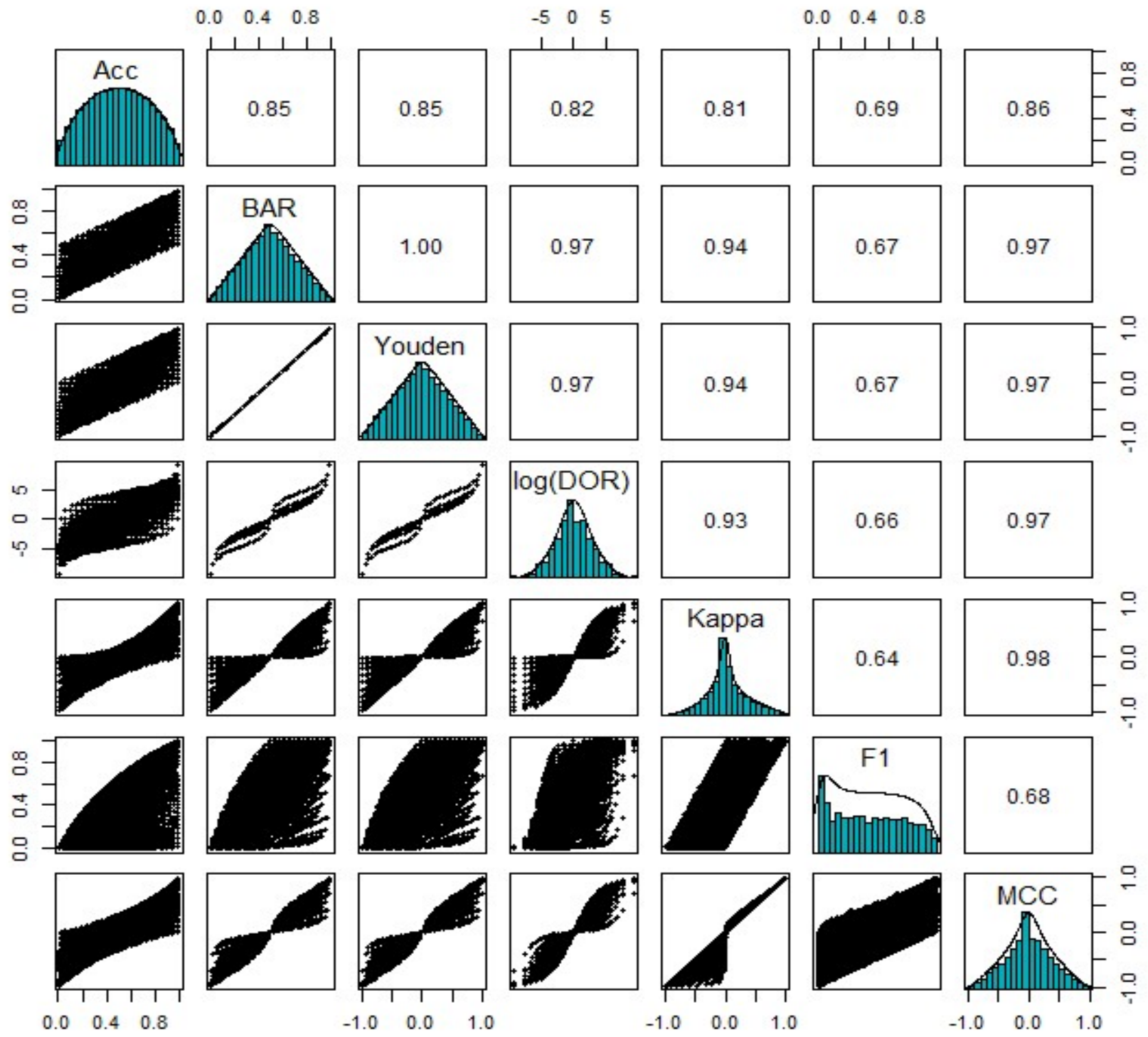

Figure A4 presents the same results as Figure A3, but excluding combinations that correspond to models that are worse than the null model (i.e. BAR<0.5). The correlations are lower, but the overall picture is the same.

Figure A4. Histograms, scatter plots and Spearman correlations for 4851 combinations of prevalence, sensitivity, and specificity for which BAR≥0.5.

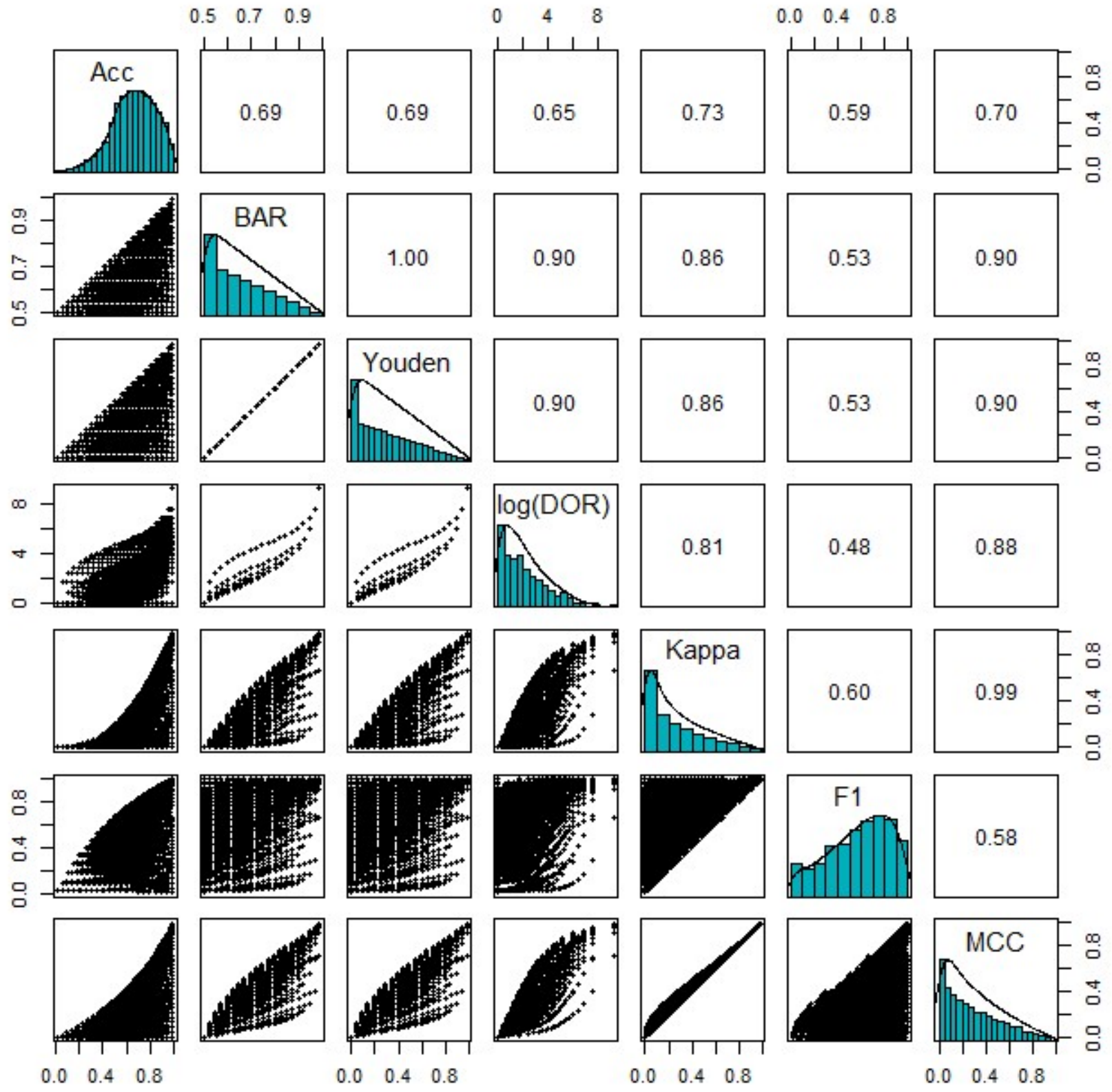

Figure A5 shows results for 231 combinations that correspond to models that are at least as good as a null model and for which the outcome is balanced (prevalence = 0.5). Now, classification accuracy correlates near perfectly with BAR/Youden, Kappa, and MCC (correlation ≥0.99), and to a slightly lesser extent with DOR (0.90). F1 also correlates relatively well with other measures (0.76-0.84).

Figure A5. Histograms, scatter plots and Spearman correlations for 231 combinations of sensitivity, and specificity for which BAR≥0.5 and prevalence is 0.5.

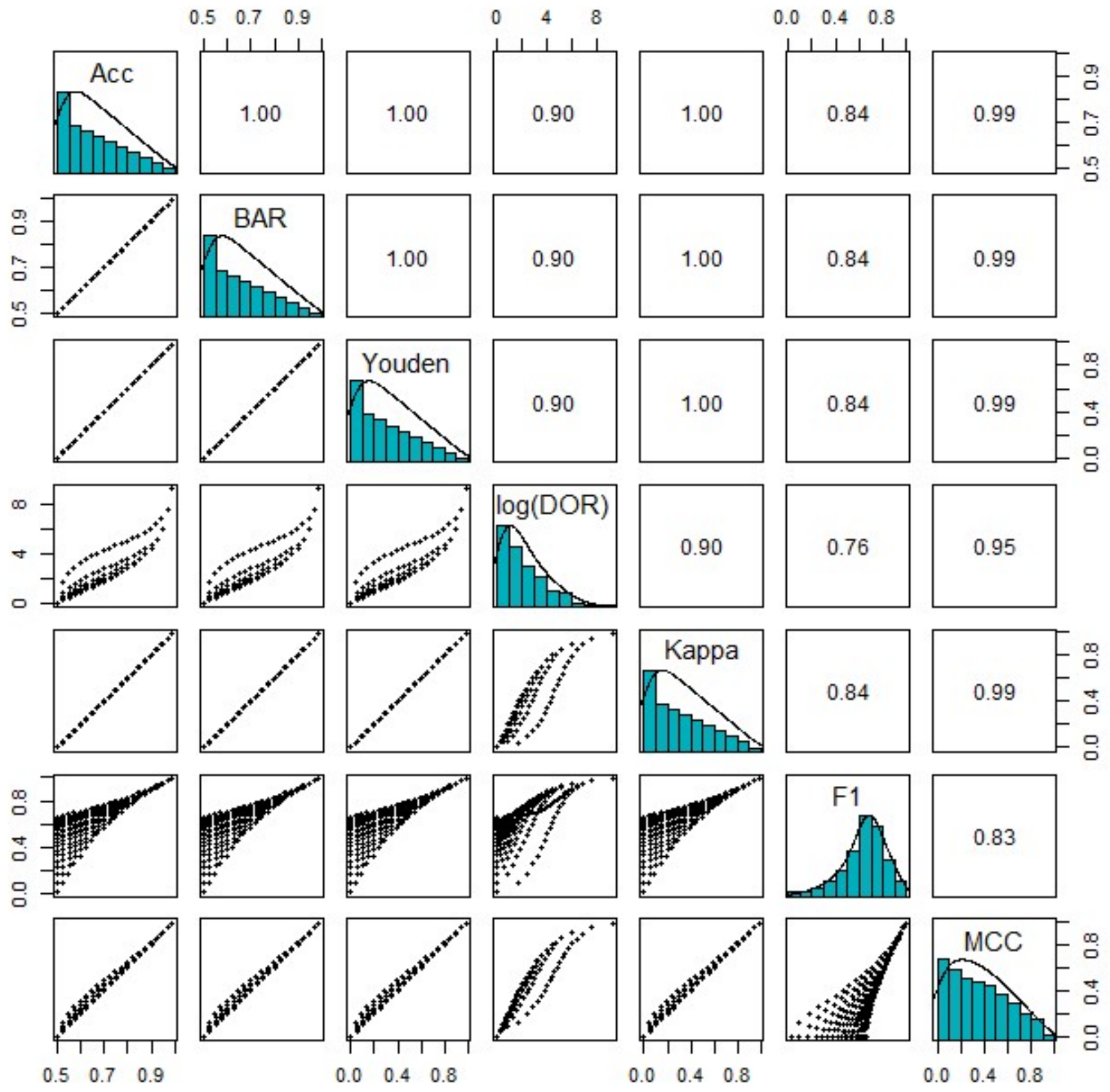

Finally, Figure A6 presents the measures by prevalence for a situation with high sensitivity and low specificity (0.9 and 0.3), a situation with moderate sensitivity and specificity (both 0.6), and a situation with low sensitivity and high specificity (0.3 and 0.9). The results show that (1) BAR/Youden and DOR are independent from prevalence, (2) kappa and MCC are lower at very low or high prevalence, (3) F1 increases with prevalence, and (4) classification accuracy increases with prevalence when sensitivity>specificity, decreases with prevalence when sensitivity<specificity, and is independent from prevalence when sensitivity=specificity.

Figure A6. Summary measures for classification by prevalence.

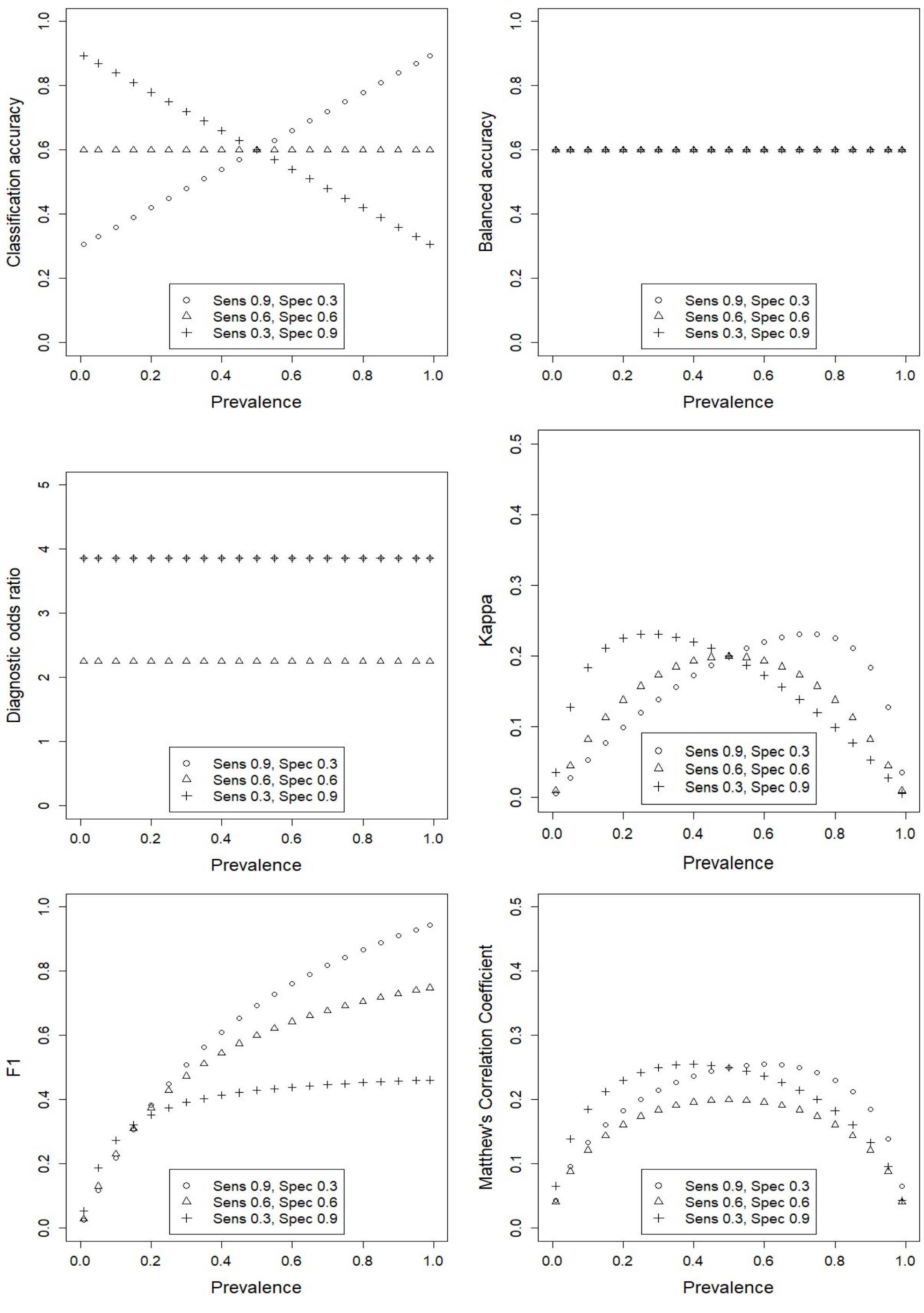